\documentclass[aps,amsmath,amssymb]{revtex4-2}
\usepackage{txfonts}
\usepackage{algorithmic}
\usepackage{algorithm}
\usepackage{graphicx}
\usepackage{bm}
\usepackage{url}
\usepackage{color}
\usepackage{svg}

\newcommand{\Var}{\operatorname{Var}}
\newcommand{\Corr}{\operatorname{Corr}}
\newcommand{\argmin}{\operatorname*{arg\,min}}
\newcommand{\argmax}{\operatorname*{arg\,max}}

\begin{document}

\title{Delay-coordinate reconstruction and conditional-moment causal diagnostics in stochastic systems}

\author{Jun Ohkubo}

\affiliation{Graduate School of Science and Engineering, Saitama University, Sakura, Saitama, 338--8570 Japan}

\begin{abstract}
Partial observation and delay-coordinate reconstruction are rooted in deterministic dynamical-systems theory, whereas there are many systems with intrinsic stochasticity. We propose a conditional-moment interpretation of delay-coordinate reconstruction for stochastic systems, in which a delay vector is used to reconstruct conditional moments of a future distribution rather than a unique future sample path. Two complementary arguments motivate this viewpoint. First, the probability density of a stochastic differential equation obeys a deterministic Fokker--Planck equation and, under certain assumptions, is represented by an infinite deterministic hierarchy of moments. Hence, a finite-moment closure suggests a Takens-like finite-dimensional approximation. Second, a discussion based on the Koopman operator theory clarifies that the time evolution of an observable in the Mori--Zwanzig formalism yields a conditional expectation in stochastic systems. Then, the orthogonal ``noise'' term in the coefficient-space Mori--Zwanzig equation vanishes in the stochastic cases; this result is consistent with the moment-based argument. As an application of this stochastic delay-reconstruction viewpoint, we revisit convergent cross mapping (CCM) for diagnosing certain causal relationships. Although CCM based on the embedding theorem cannot generally be applied to stochastic systems, it is possible to examine certain types of causal relationships by using conditional moments. Using coupled logistic systems with additive and multiplicative coupling mechanisms, we discuss how causal relationships are embedded in stochastic systems.
\end{abstract}

\maketitle

\section{Introduction}
\label{sec:introduction}

Partial observation creates a fundamental difficulty in analyzing complex dynamical systems. Complete observation of all state variables is often impossible in experiments; experiments typically provide only one or a few scalar time series. Partial observation generally destroys determinism or Markovianity: even when the full dynamics is deterministic or Markovian, the effective dynamics of the observed variables can depend on their history and on uncertainty associated with unresolved degrees of freedom. The Mori--Zwanzig projection-operator formalism provides a suitable tool for this loss of closure; see the original works by Mori and Zwanzig \cite{Zwanzig1961,Mori1965,Zwanzig1973} and some recent works \cite{Chorin2000,Chorin2002,Chorin2006,Gouasmi2017, teVrugt2024, Netz2024}. In the Mori--Zwanzig formalism, the reduced evolution is decomposed into an instantaneous or Markov term, a history-dependent memory term, and an orthogonal-dynamics term, often interpreted as a fluctuating or noise term. Recently, numerous data-driven approaches to the Mori--Zwanzig formalism have been developed \cite{Chorin2015,Meyer2019,Maeyama2020,Gonzalez2021,Tian2021}, and several studies have discussed applications to neural networks \cite{Venturi2023,Gupta2025}.

A complementary viewpoint on the Mori--Zwanzig formalism is provided by the Koopman operator theory. Rather than evolving the state itself, the Koopman operator theory yields a time evolution of observable functions of the state by a linear operator, even when the underlying state dynamics is nonlinear \cite{Koopman1931,Mezic2005}. This operator-theoretic viewpoint underlies dynamic mode decomposition (DMD) \cite{Rowley2009,Schmid2010}, extended dynamic mode decomposition (EDMD) \cite{Williams2015,Klus2016,Korda2018}, Hankel methods \cite{Brunton2017}, and many related data-driven algorithms; see reviews in \cite{Budisic2012,Mezic2013,Rowley2017,Mauroy2020,Otto2021,Mezic2021,Brunton2022,Schmid2022,Klus2024,Ghosh2024,Duran-Siguenza2025}. Several recent studies have connected the Mori--Zwanzig formalism with modern developments in the Koopman operator theory \cite{Lin2021,Lin2021a,Lin2023}. The relation between Koopman learning and Mori--Zwanzig reduction has become increasingly explicit. The data-driven Mori--Zwanzig formalism shows that a generalized Langevin representation is regarded as a systematic extension of a memoryless Koopman approximation and that Markov and memory operators are estimated from reduced time series.

The connection between the Koopman operator theory and the Mori--Zwanzig formalism leads to further developments. One of the remarkable features of the Koopman operator theory is its application to stochastic systems \cite{Williams2015,Crnjaric-Zic2020,Wanner2022}. As mentioned above, since the relationship between the Koopman operator theory and the Mori--Zwanzig formalism has been discussed in deterministic systems, it is natural to extend that discussion to stochastic systems; the previous work in \cite{Ohkubo2025} reconsidered this relation for intrinsically stochastic systems. The key observation is that one must distinguish the state space from the function space, especially in stochastic discussions. In a deterministic system, the Koopman evolution of a coordinate observable is identified with the uniquely evolved coordinate because a point initial condition remains a point under the deterministic flow. In a stochastic system, however, the same observable evolves into a function that returns a conditional expectation. Then, the orthogonal term, i.e., the noise term, of the Mori--Zwanzig equation vanishes for a coordinate observable even though the underlying state process remains stochastic.

The Mori--Zwanzig formalism yields a memory term that draws on past coordinate states, and a similar idea of past information is also discussed in the Koopman operator theory. Specifically, the delay-coordinate embedding enriches the observable space and leads to the clarification of the spectral structure \cite{Arbabi2017,Brunton2017,Clainche2017,Das2019,Kamb2020}. Even setting aside the Koopman operator theory, many discussions on delay-coordinate embedding in dynamical systems have emerged to date. In particular, Takens' embedding theorem is fundamental in dynamical systems theory and allows us to reconstruct a complex system's full state from a single observed variable; under generic smoothness assumptions, a sufficiently long scalar delay vector embeds a finite-dimensional deterministic state manifold \cite{Takens1981}. Note that Takens' embedding theorem has also been discussed in the context of stochastic systems \cite{Stark1997,Stark2003}, and more recently, measure-theoretic time-delay embedding has formulated reconstruction directly at the level of probability measures \cite{Botvinick-Greenhouse2025}. In the measure-theoretic time-delay embedding, one should distinguish reconstruction of a sample-path point from reconstruction of a distributional object. However, there has not been sufficient discussion regarding its relationship to the Mori--Zwanzig formalism and data-driven techniques such as EDMD.

The present paper focuses on partial observation and delay-coordinate embedding in stochastic systems. In particular, we clarify the relationship between Takens' theorem and delay-coordinate embedding in stochastic systems through a discussion similar to that of the Mori--Zwanzig formalism in \cite{Ohkubo2025}; the Mori--Zwanzig formalism replaces point reconstruction in deterministic cases with \emph{conditional-moment reconstruction} in stochastic cases. In intrinsically stochastic systems, information obtained from delay-coordinate embedding generally corresponds not to a unique point but to the conditional distribution of hidden states or future states, and the time-evolution equation for the moments of that distribution becomes deterministic. We will see that this argument is consistent with the Mori--Zwanzig formalism and that higher-order nonlinear dictionary functions are indispensable to it. Furthermore, we apply embeddings in probabilistic systems to problems in causal inference. While further statistical discussion is necessary to establish a practical method for causal inference, the purpose here is to demonstrate the applicability of the delay-coordinate reconstruction in probabilistic systems. As discussed later, a method known as convergent cross mapping (CCM) has been proposed for estimating causal relationships from time-series data of nonlinear and deterministic dynamical systems \cite{Sugihara2012}. Standard CCM is derived under deterministic embedding assumptions, which are generally violated by intrinsically stochastic dynamics. Then, we discuss a CCM-like method utilizing delay-coordinate embedding, which will open up possibilities for causal inference based on the embedding strategy in stochastic systems.

This article is organized as follows. Section~\ref{sec:theory} first reviews deterministic delay-coordinate embedding and then gives a moment-based, explicitly non-rigorous motivation for stochastic delay-coordinate reconstruction. We next discuss the stochastic Mori--Zwanzig formalism in stochastic systems in section~\ref{sec:MZ}, emphasizing the difference between state-space dynamics and observable-function dynamics, and derive the discrete-time conditional-moment viewpoint used in numerical analysis. Additive and multiplicative coupled logistic maps then provide a direct numerical validation. Section~\ref{sec:ccm} discusses causal inference problems as an application of the embedding in stochastic systems; we first review CCM and reformulate cross mapping as estimation of time-shifted conditional moments. Section~\ref{sec:summary} summarizes the conclusions, limitations, and directions for further work.

\section{Delay-coordinate reconstruction: background and stochastic interpretation}
\label{sec:theory}

We briefly review the embedding theorem for deterministic systems \cite{Takens1981}. We then discuss an analog of the embedding proposition for stochastic systems.

\subsection{Deterministic delay-coordinate embedding}

Consider a system of ordinary differential equations to model deterministic processes. The system state at time $t$, $\bm{x}_t \in \mathbb{R}^d$, obeys the following equation:
\begin{align}
\frac{d}{dt} \bm{x}_t = \bm{a}(\bm{x}_t).
\end{align}
Suppose that one cannot observe the full state $\bm{x}_t$, and only a scalar measurement is available:
\begin{align}
 x_{t}^{\mathrm{obs}} = h(\bm{x}_t),\qquad h:\mathbb{R}^d \to \mathbb{R},
 \label{eq:scalar_measurement}
\end{align}
where $h$ is a scalar observable function. Of course, the trajectory for the full state $\{\bm{x}_t\}$ contains all the information about the underlying system, but it is sometimes difficult to directly access the state. Instead, only a partial observation is available. For example, we might only be able to observe the first variable $x_{1,t}$, and $h(\bm{x}_t) = x_{1,t}$ in this case.

As an example, let us consider a two-variable system with $x_{1,t}$ and $x_{2,t}$, and $x_{2,t}$ is unobserved. Naively, since there is uncertainty about the initial condition for $x_{2,t}$, one may consider that the prediction for the next time step remains uncertain even if the system is deterministic. However, by utilizing the past values of $x_{1,t}$, it is actually possible to compensate for the effect of $x_{2,t}$ being unobserved.

The central question of delay-coordinate reconstruction is whether a finite list of past scalar measurements can serve as coordinates for the unobserved state. For a sampling interval $\tau > 0$, define the delay-coordinate,
\begin{align}
 \mathcal{D}_{E,\tau}(\bm{x}_t) =
 \left(
 h\left(\bm{x}_t\right), h\left(\Phi^{-\tau}(\bm{x}_t)\right),
 \ldots,
 h\left(\Phi^{-(d_\mathrm{E}-1)\tau}(\bm{x}_t)\right)
 \right)^{\top} \in \mathbb{R}^{d_\mathrm{E}},
 \label{eq:takens_delay_map}
\end{align}
where $\Phi^{-\tau}(\bm{x}_t)$ is a function that returns the state shifted back by $\tau$. The embedding theorem states that the delay-coordinate $\mathcal{D}_{E,\tau}(\bm{x}_t)$ is an embedding of the underlying manifold into delay-coordinate space, so the dynamics is conjugated to its image under the delay map under the theorem's assumptions. For mathematically precise statements, see \cite{Takens1981,Botvinick-Greenhouse2025}. Takens provided the classical generic result: for a compact $d$-dimensional manifold, a generic delay map into $\mathbb{R}^{2m+1}$ is an embedding, where $m$ is the dimension of the embedded manifold \cite{Takens1981}. Regarding the dimension required for embedding, subsequent advances in research have led to the discovery of tighter bounds \cite{Sauer1991,Baranski2020}.

\begin{figure}[t]
\centering
\includegraphics[width=120mm]{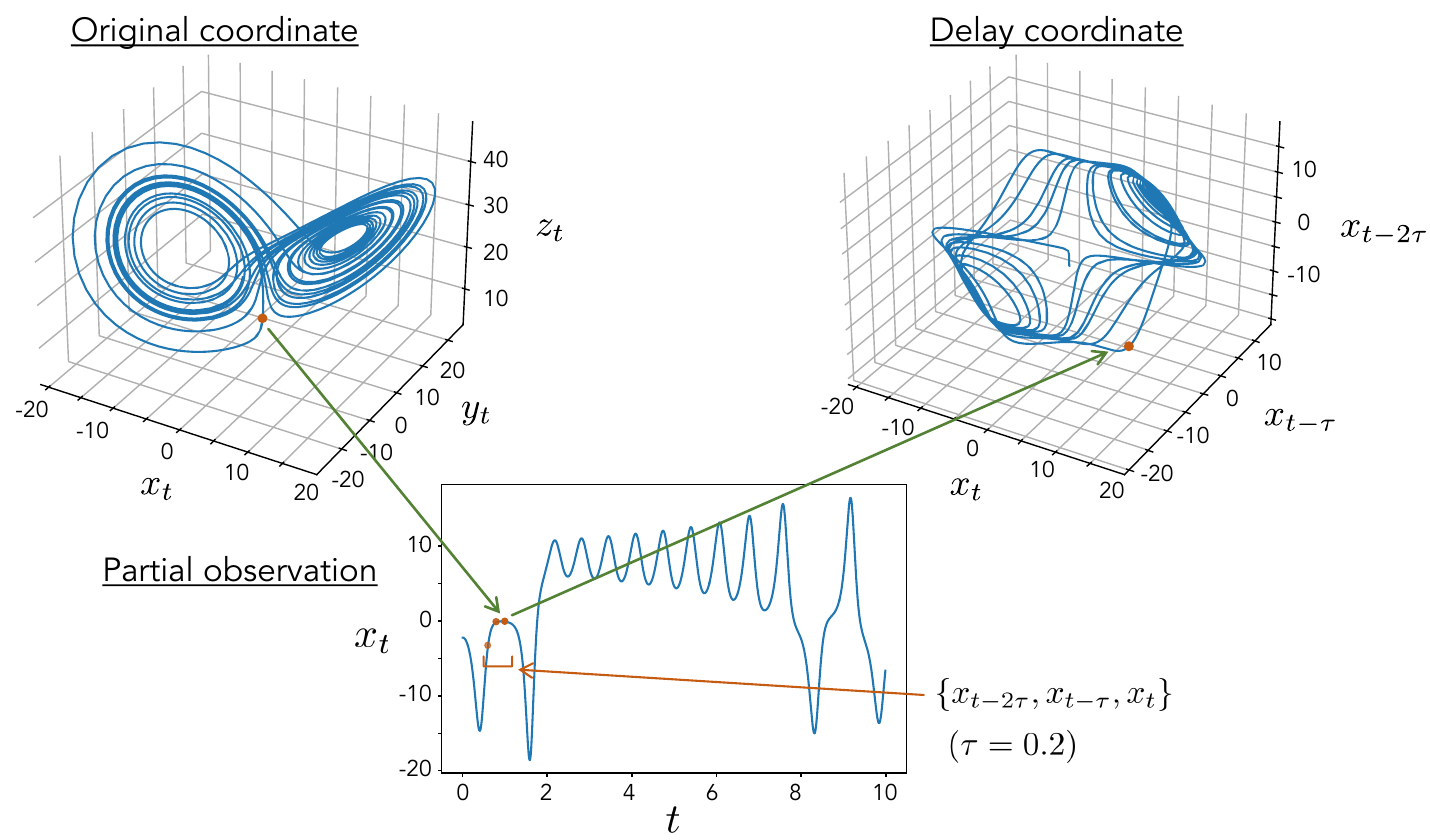}
\caption{An illustrative example of deterministic delay-coordinate reconstruction. A trajectory in the original three-dimensional state space is only partially observed through the scalar time series $x_t$.  A delay vector, here illustrated by $(x_t,x_{t-\tau},x_{t-2\tau})$ with $\tau=0.2$, can provide a geometrically equivalent coordinate representation under certain conditions.}
\label{fig:takens}
\end{figure}

Figure~\ref{fig:takens} illustrates the concept of the delay-coordinate embedding; these trajectory data were generated from the Lorenz system \cite{Lorenz1963}. In the original coordinate $(x_t,y_t,z_t)$, we see an attractor. By using the partial observation results for $x$, the delay-coordinate $(x_t, x_{t-\tau}, x_{t-2\tau})^{\top}$ also yields an attractor. Note that the shape of the attractors seems different from the original one. However, there is a certain type of one-to-one correspondence between these two attractors. Hence, analyzing the attractor in the delay-coordinate space yields valuable insights into the original system.

\subsection{Embedding in stochastic systems}
\label{sec:takens_moments}

In \cite{Botvinick-Greenhouse2025}, measure-theoretic time-delay embedding was discussed, which is applicable to spaces of probability measures. In the following, we intuitively discuss moment-based heuristic embedding for stochastic systems from a slightly different perspective.

Consider a $d$-dimensional It\^o stochastic differential equation (SDE),
\begin{align}
 d\bm{X}_t=\bm{a}(\bm{X}_t) dt + B(\bm{X}_t) d\bm{W}_t,
 \label{eq:sde_general}
\end{align}
where $\bm{W}_t$ is an $r$-dimensional Wiener process, $\bm{a}(\bm{x})$ is a coefficient vector function for the drift, and $B(\bm{x})$ is a coefficient matrix function for the diffusion. Let $D(\bm{x})$ be the diffusion matrix defined as follows:
\begin{align}
 D(\bm{x})=B(\bm{x})B(\bm{x})^{\top}.
 \label{eq:diffusion_matrix}
\end{align}
If the transition law admits a sufficiently regular density, the density $p(\bm{x},t)$ evolves according to the Fokker--Planck equation \cite{Risken1996,Gardiner2009}
\begin{align}
 \frac{\partial p}{\partial t}
=
\mathcal{L}^{\dagger} p
=
-\sum_i \frac{\partial}{\partial x_i} \left(a_i p \right)
 + \frac{1}{2} \sum_{i,j} \frac{\partial^2}{\partial x_i \partial x_j}
 \left(D_{ij} p\right).
 \label{eq:fokker_planck}
\end{align}
Thus, although individual sample trajectories are random, the time evolution of the probability density $p(\bm{x},t)$ is deterministic once the initial density is specified. 

It is also possible to discuss the time evolution of stochastic systems from the viewpoint of moments. For a multi-index $\bm{\alpha}=(\alpha_1, \dots, \alpha_d) \in \mathbb{N}_0^d$, define
\begin{align}
m_{\bm{\alpha},t} 
= \mathbb{E}[\bm{X}_t^{\bm\alpha}]
=\int d\bm{x} \, \bm{x}^{\bm{\alpha}}p(\bm{x},t), 
 \qquad 
\bm{x}^{\bm{\alpha}}
=\prod_i x_i^{\alpha_i}.
 \label{eq:moment_def}
\end{align}
Then, the backward Kolmogorov generator $\mathcal{L}$ is given as
\begin{align}
\mathcal{L} f
=
\sum_i a_i(\bm{x})\partial_i f
+\frac{1}{2} \sum_{i,j} D_{ij}(\bm{x}) \partial_i\partial_j f,
 \label{eq:backward_generator}
\end{align}
and It\^o's formula gives the following time-evolution equations for the moments \cite{Gardiner2009}:
\begin{align}
 \frac{d}{dt}m_{\bm{\alpha}, t}
&= \mathbb{E} [\mathcal{L} \bm{X}_t^{\bm{\alpha}}] \nonumber \\
&= \sum_i \alpha_i \, \mathbb{E} \left[
 \bm{X}_t^{\bm{\alpha}-\bm{e}_i} a_i(\bm{X}_t)\right]
+\frac{1}{2}\sum_{i,j} \alpha_i(\alpha_j - \delta_{ij})
\, \mathbb{E} \left[
 \bm{X}_t^{\bm{\alpha}-\bm{e}_i-\bm{e}_j} D_{ij} (\bm{X}_t)\right],
 \label{eq:moment_hierarchy}
\end{align}
where $\bm{e}_i$ is the standard basis vector. Note that taking the expectations in \eqref{eq:moment_hierarchy} eliminates explicit noise terms. For polynomial drift and diffusion coefficient functions, each right-hand side is a linear combination of moments, but nonlinear coefficient functions generally connect low-order moments to higher-order moments. Hence, the exact moment representation is an infinite deterministic hierarchy, i.e.,
\begin{align}
 \frac{d\bm{M}_{\infty}}{dt}=\bm F(\bm{M}_{\infty}),
 \qquad
 \bm{M}_{\infty}=\{m_{\bm{\alpha}}\}_{\bm{\alpha}\in\mathbb{N}_0^d}.
 \label{eq:infinite_moments}
\end{align}
Note that $\bm{M}_{\infty}$ is an infinite-dimensional vector, and the function $F$ is a vector-valued function that returns an infinite-dimensional vector. 

From the discussion above, one would expect that it is possible to reconstruct the probability density function $p(\bm{x},t)$ if all of its moments are known. However, the existence of moments of all orders does not by itself imply that the probability density function is uniquely determined. The reconstruction is true under a certain condition, and we therefore restrict the present heuristic argument to moment-determinate distributions \cite{Lin2017}. For example, in one dimension,
\begin{align}
 \sum_{k=1}^{\infty}m_{2k}^{-1/(2k)} = \infty
 \label{eq:carleman}
\end{align}
is a sufficient condition for determinacy. Under such assumptions, one may schematically regard
\begin{align}
 p(\bm{x},t)
 \quad\longleftrightarrow\quad
 \{m_{\bm\alpha}(t)\}_{\bm\alpha}
 \label{eq:density_moment_equivalence}
\end{align}
as an information-preserving representation at the infinite-dimensional level.

The infinite hierarchy~\eqref{eq:infinite_moments} is not a finite-dimensional smooth dynamical system, and therefore the ordinary Takens' theorem cannot be applied to it directly. A practical calculation retains only finitely many moments,
\begin{align}
 \bm{M}_K=\left(m_{\bm{\alpha}}\right)_{|\bm{\alpha}|\le K},
 \quad
 \bm{M}_{>K}=\left(m_{\bm{\alpha}}\right)_{|\bm{\alpha}|> K},
\quad
 \frac{d\bm{M}_K}{dt}=\bm{F}_K(\bm{M}_K,\bm{M}_{>K}).
 \label{eq:finite_moment_unclosed}
\end{align}
A moment closure or data-driven approximation produces an approximate finite deterministic system
\begin{align}
 \frac{d\bm{M}_K}{dt}=\widetilde{\bm{F}}_K(\bm{M}_K).
 \label{eq:finite_moment_closed}
\end{align}
If this approximate system satisfies the appropriate conditions for the embedding theorem, it is possible to apply the deterministic delay-map construction to a scalar observable $h_K(\bm{M}_K)$. Formally,
\begin{align}
 \bm{M}_K
 \mapsto
 \left(
 h_K(\bm{M}_K),
 h_K(\Phi_K^{-\tau}\bm{M}_K),\ldots,
 h_K(\Phi_K^{-(d_\mathrm{E}-1)\tau}\bm{M}_K)
 \right)^{\top},
 \label{eq:moment_delay_map}
\end{align}
is then an ordinary deterministic embedding problem for the \emph{closed approximate moment dynamics} with a sufficiently large $d_\mathrm{E}$.

Hence, one might expect the following relationship:
\begin{align*}
p(\bm{x},t)
\longleftrightarrow \bm{M}_{\infty}(t)
 \longleftrightarrow \bm{M}_K(t)
 \longleftrightarrow \textrm{finite-dimensional delay representation}.
\end{align*}
Note that strictly speaking, the second double-headed arrow does not hold exactly; rather, it holds only approximately. The discussion here is intended to provide an intuitive understanding of how moments, i.e., expected values, arise in the embedding of stochastic systems in the following discussions. Rigorous stochastic embedding results impose different hypotheses \cite{Stark1997,Stark2003}, while measure-theoretic time-delay embedding works directly with measures and pushforwards rather than with a finite moment closure \cite{Botvinick-Greenhouse2025}. The present paper does not attempt to prove equivalence among these formulations.

Nevertheless, the moment argument suggests an important change of viewpoint: in a stochastic system, the natural object is the conditional law, and its conditional moments provide deterministic summaries of that law. Although a mathematically rigorous proof is beyond the scope of the present paper, in the subsequent section, we will see the same conclusion from a different route, i.e., the Koopman operator theory and the Mori--Zwanzig formalism, without assuming a finite moment closure.

\section{Mori--Zwanzig projection in stochastic Koopman observable space}
\label{sec:MZ}

Here, we briefly review the discussion in \cite{Ohkubo2025} and discuss it from a different perspective. Then, we discuss how to connect an operator formalism to a partially observed dataset.

\subsection{The generalized Langevin equation in deterministic systems}

Let $\mathcal{P}$ be a projection onto the observable function space and let $\mathcal{Q}= I-\mathcal{P}$. For a generator $\mathcal{L}$ and a function $g$, the Dyson identity yields the following decomposition \cite{Lin2021a}
\begin{align}
 \frac{d}{dt}e^{t \mathcal{L}}g
 = e^{t\mathcal{L}}\mathcal{P}\mathcal{L} g
 + \int_0^t ds \, e^{(t-s)\mathcal{L}}\mathcal{P}\mathcal{L} e^{s\mathcal{Q}\mathcal{L}}\mathcal{Q}\mathcal{L} g
 + e^{t\mathcal{Q}\mathcal{L}} \mathcal{Q} \mathcal{L} g.
 \label{eq:standard_mz_identity}
\end{align}
The three terms on the right-hand side are conventionally interpreted as an instantaneous resolved contribution, a memory contribution, and an orthogonal-dynamics contribution, respectively. Note that the third orthogonal term is sometimes called ``noise term,'' but it does not correspond generally to simple white Gaussian noise; it is generated by the unobserved dynamics. 

Here, consider a deterministic system in which we observe only a variable $x$, and let $\chi(\bm{x})$ be the observable function, i.e., $\chi(\bm{x}) = x$. Then, $e^{t \mathcal{L}} \chi$ yields the time evolution of the observable function which returns the state variable $x_t$ at time $t$. In a deterministic system, if the initial state is given by a Dirac delta function, the state after time evolution is also given by a Dirac delta function. Therefore, as discussed in \cite{Lin2021a,Ohkubo2025}, the time evolution of observable functions is regarded as the time evolution of the state variables. Hence, the following generalized Langevin equation for a certain linear representation is derived:
\begin{align}
 \frac{d}{dt} x_t
 = M_{\mathrm{trans}} x_t
 + \int_0^t ds \, K_{\mathrm{mem}} (t-s) x_s 
 + (\textrm{noise term}).
 \label{eq:generalized_Langevin}
\end{align}
For a detailed discussion, see \cite{Ohkubo2025}. The crucial point in the present paper is that this argument does not hold for stochastic systems; instead, it is necessary to conduct the discussion using the time-evolution equation for the observable functions rather than that for the state variables.

\subsection{Koopman operator theory}

For details of the Koopman operator theory and discussions on a dictionary function space, see, for example, \cite{Williams2015}.

First, we consider a deterministic system. Let $\mathcal{F}$ be a suitable space of scalar observables $\phi:\mathcal{M} \to \mathbb{R}$. For a deterministic flow $\Phi^t$, the Koopman operator is the composition operator
\begin{align}
 (\mathcal{K}_t\phi)(\bm{x})=\phi\left(\Phi^t(\bm{x})\right),
 \label{eq:det_koopman_composition}
\end{align}
where $\Phi^t$ is the state-space flow, i.e., $\Phi^{\tau}(\bm{x}_t) = \bm{x}_{t+\tau}$. The Koopman operator satisfies the following property $\mathcal{K}_{t+s}=\mathcal{K}_t\mathcal{K}_s$ and is linear in the observable,
\begin{align}
 \mathcal{K}_t( a \phi_1 + b\phi_2)
=a\mathcal{K}_t \phi_1 + b \mathcal{K}_t\phi_2,
 \label{eq:koopman_linearity}
\end{align}
even when $\Phi^t$ is nonlinear. 

Here, note that a state value and a coordinate observable are different mathematical objects. If the state is $\bm{x}=(x_1,\dots,x_d)^{\top}$, the coordinate observable $\chi_i \in \mathcal{F}$ is the function
\begin{align}
 \chi_i(\bm{x})=x_i.
 \label{eq:coordinate_observable}
\end{align}
Then, for the coordinate observable,
\begin{align}
 (\mathcal{K}_t \, \chi_i)(\bm{x})=\chi_i\left(\Phi^t(\bm{x})\right)
=\left[\Phi^t (\bm{x}) \right]_i,
 \label{eq:det_coordinate_koopman}
\end{align}
and deterministic function evolution and deterministic state evolution are identified after evaluation at the initial state. 

Second, consider a stochastic system; we here consider a Markov process obeying \eqref{eq:sde_general}. Then, the Koopman operator yields the conditional expectation \cite{Williams2015}:
\begin{align}
 (\mathcal{K}_t\phi)(\bm{x})
 =\mathbb{E}\left[ \phi(\bm{X}_t) \mid \bm{X}_0=\bm{x}\right]
 =\int d\bm{x}' \,  \phi(\bm{x}') \, p_t(\bm{x}' \mid \bm{x}),
 \label{eq:stoch_koopman}
\end{align}
where $p_t(\bm{x}' \mid \bm{x})$ is the transition density. For the SDE in \eqref{eq:sde_general}, we have
\begin{align}
 \mathcal{K}_t=e^{t \mathcal{L}},
 \label{eq:koopman_semigroup_generator}
\end{align}
with the backward generator $\mathcal{L}$ in \eqref{eq:backward_generator}. The adjoint semigroup $e^{t\mathcal{L}^\dagger}$ evolves the corresponding probability density, and the duality relation gives
\begin{align}
 \int d\bm{x} \, \phi(\bm{x}) \, \left[e^{t\mathcal{L}^\dagger} p_0 \right](\bm{x})
 = \int d\bm{x} \left[e^{t\mathcal{L}}\phi \right](\bm{x}) \, p_0(\bm{x}).
 \label{eq:forward_backward_duality}
\end{align}
Taking $p_0(\bm{x})=\delta(\bm{x}-\bm{x}_0)$ recovers the same form as \eqref{eq:stoch_koopman}. For details, see~\cite{Ohkubo2025}.

Here, choose the observable as $\phi(\bm{x}) = x_i^{m}$. Then, we have
\begin{align}
 (\mathcal{K}_t \phi)(\bm{x}_0)
 =\mathbb{E}[X_{i,t}^{m} \mid \bm{X}_0=\bm{x}_0],
 \label{eq:coordinate_condmoment_cont}
\end{align}
which is the $m$-th conditional moment of $X_{i,t}$ given $\bm{X}_0=\bm{x}_0$.

Next, we consider a concrete representation of the Koopman operator. To this end, we assume that the function space $\mathcal{F}$ is spanned by dictionary functions $\{\psi_i(\bm{x})\}$. In general, the number of dictionary functions is infinite. Denoting
\begin{align}
\bm{\psi}_{\mathrm{inf}}(\bm{x}) = \bigl(\psi_1(\bm{x}),\psi_2(\bm{x}), \ldots\bigr)^{\top},
\label{eq:dictionary_infinite}
\end{align}
and assuming the dictionary spans an invariant function space $\mathcal{F}$, $\mathcal{K}_t \psi_i$ is expanded as
\begin{align}
\mathcal{K}_t \psi_i(\bm{x}) = \sum_{j} c_{ij}(t) \psi_{j}(\bm{x}),
\label{eq:dictionary_span}
\end{align}
where $\{c_{ij}(t)\}$ is a set of the expansion coefficients. Here, for clarity, the time variable $t$ is denoted as an argument rather than as a subscript. Note that $\mathcal{K}_t \psi_i(\bm{x})$ is also a function in $\mathcal{F}$, and it is possible to express it as a linear combination of $\{\psi_i(\bm{x})\}$. Then, as discussed in \cite{Ohkubo2019}, the generator $\mathcal{L}$ leads to a coupled ordinary differential equation for the expansion coefficients $\{c_{ij}(t)\}$.

In what follows, we discuss the Mori--Zwanzig formalism using the expansion coefficients $\{c_{ij}(t)\}$. It should be noted that these expansion coefficients are not state variables, but they are discussed strictly within the context of the function space $\mathcal{F}$.

\subsection{Projections in coefficient space and disappearance of noise term}

Similar to \eqref{eq:dictionary_infinite}, we introduce dictionary functions. The dictionary $\bm{\psi}(\bm{x})$ contains the observed coordinate functions as well as additional unresolved dictionary functions. Then, we partition it as
\begin{align}
 \bm{\psi}(\bm{x}) 
= \left(
\Bigl(\bm{\psi}_{\mathcal{O}}(\bm{x})\Bigr)^{\top},
\Bigl(\bm{\psi}_{\overline{\mathcal{O}}}(\bm{x})\Bigr)^{\top} \right)^{\top},
 \label{eq:dictionary_partition}
\end{align}
where $\mathcal{O}$ denotes the retained observable subspace and $\overline{\mathcal{O}}$ corresponds to its complement inside the dictionary. An evolving observable is expanded as
\begin{align}
 \phi(\bm{x},t)
=
\bm{c}_{\mathcal{O}}(t)^{\top}\bm{\psi}_{\mathcal{O}}(\bm{x})
+ \bm{c}_{\overline{\mathcal{O}}}(t)^{\top}\bm{\psi}_{\overline{\mathcal{O}}}(\bm{x})
 \label{eq:function_basis_split}
\end{align}
Note that ~\eqref{eq:function_basis_split} is an equation in the function space. The vectors $\bm{c}_{\mathcal{O}}(t)$ and $\bm{c}_{\overline{\mathcal{O}}}(t)$ are expansion coefficients of the corresponding observable; they are not the observed and unresolved components of a stochastic \emph{state} vector.

In the following discussions, we employ monomial basis functions since they are useful for discussing the conditional moment aspects. Hence, a dictionary function has the following form:
\begin{align}
\psi_j(\bm{x}) = \chi^{\bm{\alpha}_j}(\bm{x})
= \prod_{k} x_k^{\alpha_{j,k}}.
\end{align}
For example, when observing only the first variable $x_1$, we define
\begin{align}
\bm{\psi}_{\mathcal{O}}(\bm{x})
= \bigl(1, x_1, x_1^2, x_1^3, \dots \bigr)^{\top}, 
\end{align}
and other monomial basis functions are in $\bm{\psi}_{\overline{\mathcal{O}}}(\bm{x})$.

In order to employ orthogonality with respect to the monomial basis functions, we formally introduce the following dual basis; 
\begin{align}
\widetilde{\psi}_j(\bm{x})
= \frac{1}{\bm{\alpha}_j!} \prod_{k} \delta(x_{k}) \left( \frac{\partial}{\partial x_k}\right)^{\alpha_{j,k}} (\cdot),
\label{eq:dual_basis}
\end{align}
where $\bm{\alpha}_j! = \prod_k \alpha_{j,k}!$. Then, we define the inner product as follows:
\begin{align}
\langle 
\widetilde{\psi}_j,\, \psi_{j'} 
\rangle
= \int \prod_k dx_k \,\delta(x_k) \left( \frac{\partial}{\partial x_k}\right)^{\alpha_{j,k}} (x_{k})^{\alpha_{j',k}}
= \delta_{j,j'},
\label{eq:orthogonal}
\end{align}
where $\delta_{j,j'}$ is the Kronecker delta function.

As discussed above, we consider the time evolution of the expansion coefficients $\bm{c}_{\mathcal{O}}(t)$ and $\bm{c}_{\overline{\mathcal{O}}}(t)$ instead of the observable function. That is, from the Koopman operator theory, the time evolution of the observable function $\phi(\bm{x},t)$ is given by
\begin{align}
\frac{d}{dt} \phi(\bm{x}, t) = \mathcal{L} \phi(\bm{x}, t),
\end{align}
and the expansion in \eqref{eq:function_basis_split} leads to
\begin{align}
\frac{d}{dt} 
\left(\bm{c}_{\mathcal{O}}(t)^{\top}\bm{\psi}_{\mathcal{O}}(\bm{x})
+ \bm{c}_{\overline{\mathcal{O}}}(t)^{\top}\bm{\psi}_{\overline{\mathcal{O}}}(\bm{x})
\right)
=  \mathcal{L} \left(\bm{c}_{\mathcal{O}}(t)^{\top}\bm{\psi}_{\mathcal{O}}(\bm{x})
+ \bm{c}_{\overline{\mathcal{O}}}(t)^{\top}\bm{\psi}_{\overline{\mathcal{O}}}(\bm{x})
\right).
\end{align}
Using the orthogonality relation in \eqref{eq:orthogonal}, we define
\begin{align}
 L_{ij} = \langle \widetilde{\psi}_i, \mathcal{L} \psi_j\rangle,
 \label{eq:gram_generator_matrices}
\end{align}
and the following time-evolution equation is obtained:
\begin{align}
 \frac{d}{dt}
 \begin{bmatrix}
\bm{c}_{\mathcal{O}}(t)\\
\bm{c}_{\overline{\mathcal{O}}} (t)
\end{bmatrix}
 =
 \begin{bmatrix}
 L_{\mathcal{O}\mathcal{O}} & L_{\mathcal{O}\overline{\mathcal{O}}}\\
 L_{\overline{\mathcal{O}}\mathcal{O}} &  L_{\overline{\mathcal{O}}\overline{\mathcal{O}}}
 \end{bmatrix}
 \begin{bmatrix}
\bm{c}_{\mathcal{O}}(t)\\
\bm{c}_{\overline{\mathcal{O}}} (t)
\end{bmatrix},
 \label{eq:block_coeff}
\end{align}
where the submatrices $ L_{\mathcal{O}\mathcal{O}}, L_{\mathcal{O}\overline{\mathcal{O}}},  L_{\overline{\mathcal{O}}\mathcal{O}},$ and $L_{\overline{\mathcal{O}}\overline{\mathcal{O}}}$ are appropriately defined in accordance with the subset of the basis functions $\bm{\psi}_{\mathcal{O}}(\bm{x})$ and $\bm{\psi}_{\overline{\mathcal{O}}}(\bm{x})$.

Then, we solve \eqref{eq:block_coeff} for $\bm{c}_{\overline{\mathcal{O}}}(t)$ implicitly;
\begin{align}
\bm{c}_{\overline{\mathcal{O}}}(t) 
= 
\int_0^{t} e^{(t-s) L_{\overline{\mathcal{O}} \overline{\mathcal{O}}}} L_{\overline{\mathcal{O}} \mathcal{O}} \bm{c}_{\mathcal{O}}(s) ds
+ e^{t L_{\overline{\mathcal{O}} \overline{\mathcal{O}}}} \bm{c}_{\overline{\mathcal{O}}}(0).
\label{eq:MZ_1}
\end{align}
Hence, inserting \eqref{eq:MZ_1} into \eqref{eq:block_coeff}, we have
\begin{align}
\frac{d}{dt} \bm{c}_{\mathcal{O}}(t)
=
L_{\mathcal{O} \mathcal{O}} \bm{c}_{\mathcal{O}}(t)
+ L_{\mathcal{O} \overline{\mathcal{O}}} 
\int_0^{t} e^{(t-s) L_{\overline{\mathcal{O}} \overline{\mathcal{O}}}} L_{\overline{\mathcal{O}} \mathcal{O}} \bm{c}_{\mathcal{O}}(s) ds
+  L_{\mathcal{O} \overline{\mathcal{O}}} e^{t L_{\overline{\mathcal{O}} \overline{\mathcal{O}}}} \bm{c}_{\overline{\mathcal{O}}}(0).
\label{eq:MZ_2}
\end{align}
Replacing $L_{\mathcal{O} \mathcal{O}}$ with $M_{\mathrm{trans}}$ and $- L_{\mathcal{O} \overline{\mathcal{O}}} e^{s L_{\overline{\mathcal{O}} \overline{\mathcal{O}}}} L_{\overline{\mathcal{O}} \mathcal{O}}$ with $K_{\mathrm{mem}}(s)$, we have the following generalized Langevin equation:
\begin{align}
\frac{d}{dt} \bm{c}_{\mathcal{O}}(t)
=
M_{\mathrm{trans}} \, \bm{c}_{\mathcal{O}}(t)
- \int_0^{t} K_{\mathrm{mem}}(t-s) \bm{c}_{\mathcal{O}}(s) ds + \bm{R}(t),
 \label{eq:mz_coeff}
\end{align}
where $\bm{R}(t) = L_{\mathcal{O} \overline{\mathcal{O}}} e^{t L_{\overline{\mathcal{O}} \overline{\mathcal{O}}}} \bm{c}_{\overline{\mathcal{O}}}(0)$. Again, we note the following fact: \eqref{eq:mz_coeff} has the formal Markov--memory--orthogonal structure of a generalized Langevin equation, but the dynamical variable is the coefficient vector of an observable function.

Next, we discuss the orthogonal term $\bm{R}(t)$ in \eqref{eq:mz_coeff}. For simplicity, assume that only the first variable is observable, i.e., $\chi_1(\bm{x})=x_1$. Then, the initial condition for $\phi(\bm{x},0)$ is given by
\begin{align}
 \phi(\bm{x},0)=\chi_1(\bm{x}) = x_1.
 \label{eq:coordinate_initial_observable}
\end{align}
That is, its expansion coefficients are one-hot: the coefficient multiplying $\chi_1$ is unity and all other coefficients, including all unresolved coefficients, vanish. Symbolically, if $i_{x_1}$ is the index of the basis function corresponding to the function $\chi_{1}$, we have
\begin{align}
 \bm{c}_{\mathcal{O}}(0) = \bm{e}_{i_{x_1}},
 \qquad
 \bm{c}_{\overline{\mathcal{O}}}(0) = \bm{0},
 \label{eq:onehot_initial}
\end{align}
where $\bm{e}_i$ is the standard basis vector.
Hence,
\begin{align}
 \bm{R}(t) = 
L_{\mathcal{O} \overline{\mathcal{O}}} 
e^{t L_{\overline{\mathcal{O}}\overline{\mathcal{O}}}} 
\bm{c}_{\overline{\mathcal{O}}}(0)
=\bm{0},
 \label{eq:orthogonal_zero}
\end{align}
and \eqref{eq:mz_coeff} reduces to
\begin{align}
\frac{d}{dt} \bm{c}_{\mathcal{O}}(t)
=
M_{\mathrm{trans}} \, \bm{c}_{\mathcal{O}}(t)
- \int_0^{t} K_{\mathrm{mem}}(t-s) \bm{c}_{\mathcal{O}}(s)  ds.
 \label{eq:mz_no_orthogonal}
\end{align}
Then, the ``noise'' term, i.e., the orthogonal term, disappears in the stochastic Koopman formulation.

Below, there are two comments on the above discussion.

First, the discussion clarified the relationship between the embedding proposition in stochastic systems and the Mori--Zwanzig projection method. Both frameworks encode information from past observations, but \eqref{eq:mz_no_orthogonal} is a coefficient-space memory equation. Note that a separate discretization/observation map is required before identifying its history dependence with a finite delay vector. Of course, \eqref{eq:mz_no_orthogonal} uses continuous-time values, so there is a gap between it and Takens' embedding theorem. Although a mathematically rigorous discussion of this point is beyond the scope of the present paper, it has at least been shown that the use of the delay-coordinate values of the observed variables enables a deterministic argument without the noise term, as in \eqref{eq:mz_no_orthogonal}.

Second, in the Mori--Zwanzig-based discussion with a linear/first-order resolved dictionary in \cite{Ohkubo2025}, the observable functions lack higher-order monomials. By contrast, there are coefficients corresponding to the higher-order monomial terms in \eqref{eq:mz_no_orthogonal}. A nonlinear function class is generally required to approximate the inverse/reconstruction map; polynomial monomials are a convenient choice. In addition, note that the above discussions do not include cross-terms involving different times. That is, the memory term in \eqref{eq:mz_no_orthogonal} refers only to time $s$; there are no coefficients corresponding to terms such as $\bm{c}_{\mathcal{O}}(s) \bm{c}_{\mathcal{O}}(s-\tau)$. Therefore, when considering higher-order dictionary functions, cross-terms involving different times are unnecessary.

\subsection{Discrete-time conditional moments from observed delay histories}
\label{sec:discrete_condmom}

To connect the operator formulation to data from partial observations, we use delays to specify the conditioning information and reconstruct the resulting conditional moments. Hence, the following discrete-time formulation is employed.

For the numerical analysis, it is convenient to work with a discrete-time Markov process; using a fixed lag time $\tau$, we set $t = 0, \tau, 2\tau, \dots$. Even in the discrete-time case, the Koopman operator is available, which  yields conditional moments after one-step time evolution:
\begin{align}
 (\mathcal{K}\phi)(\bm{x})
 =\mathbb{E}[\phi(\bm{X}_{t+\tau})\mid\bm{X}_t=\bm{x}],
 \label{eq:discrete_koopman}
\end{align}
where we denote $\mathcal{K}_{\tau}$ by $\mathcal{K}$ for notational simplicity. Choosing the monomial coordinate observable function
\begin{align}
 \phi_{j,m}(\bm{x})=x_j^m
 \label{eq:coordinate_monomial}
\end{align}
gives
\begin{align}
 (\mathcal{K}\phi_{j,m})(\bm{x})
 =\mathbb{E}[X_{j,t+\tau}^{m} \mid \bm{X}_t=\bm{x}].
 \label{eq:fullstate_condmoment}
\end{align}
This equation makes the role of the ``initial value'' particularly clear. To obtain the $m$-th conditional moment, one does not initialize the Koopman equation with an already averaged moment; if the coordinates at time $t$ are known, instead of conditioning on the moments, conditioning on those coordinate values yields the conditional expectations of the moments after the time evolution.

Suppose that only a scalar random variable $A_t$ is observed. Define a delay history as
\begin{align}
 H_t^A =(A_t, A_{t-\tau}, \dots, A_{t-q\tau}).
 \label{eq:delay_history}
\end{align}
Conditional expectation has an important variational characterization: among all square-integrable functions $g(H_t^A)$ of the available history, the conditional expectation is a solution of a least-squares problem \cite{Durrett_book}, i.e., 
\begin{align}
 \mathbb{E}[A_{t+\tau}^m \mid H_t^A]
 =\argmin_{g}
 \mathbb{E} \left[\bigl(A_{t+\tau}^m - g(H_t^A)\bigr)^2\right].
 \label{eq:conditional_expectation_projection}
\end{align}
Thus, a regression trained with squared loss has a precise stochastic target: it approximates the orthogonal projection of the future moment onto the information contained in the observed history.

\begin{figure}[t]
\centering
\includegraphics[width=120mm]{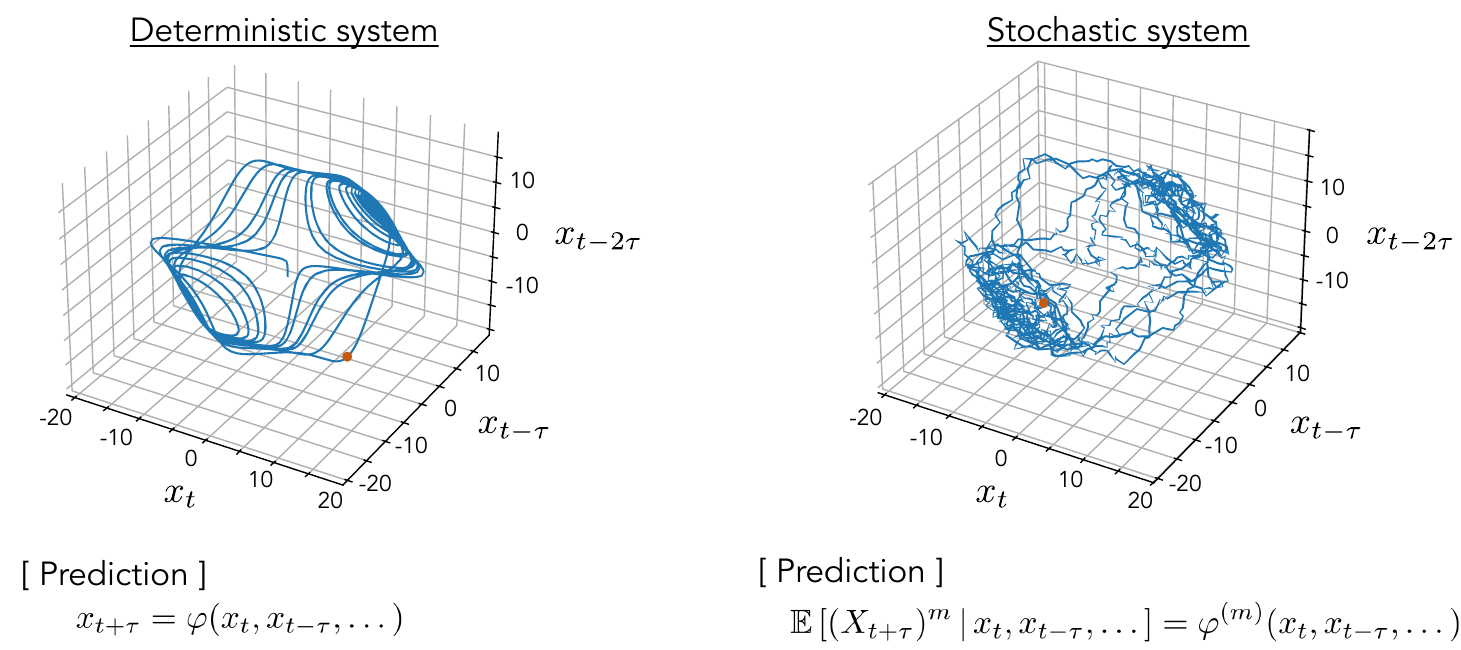}
\caption{Deterministic and stochastic interpretations of delay-coordinate prediction.  In a deterministic system, a sufficiently informative delay vector can determine a future value as $x_{t+\tau}=\varphi(x_t,x_{t-\tau},\ldots)$. In a stochastic system, the corresponding deterministic object is a conditional statistic, for example $\mathbb{E}[X_{t+\tau}^m\mid X_t,X_{t-\tau},\ldots]=\varphi^{(m)}(X_t,X_{t-\tau},\ldots)$. For example, the conditional mean describes location, while the variance, obtained from the first and second moments, quantifies stochastic spread.}
\label{fig:detstoch}
\end{figure}

Figure~\ref{fig:detstoch} summarizes a distinction between deterministic and stochastic cases. The relation to deterministic embedding is straightforward. If the dynamics is deterministic and $H_t^A$ is an injective embedding of the underlying state, then $A_{t+\tau}$ is a deterministic function of $H_t^A$ and
\begin{align}
 \Var(A_{t+\tau} \mid H_t^A)=0.
 \label{eq:deterministic_conditional_collapse}
\end{align}
Hence, the conditional law collapses to a point and $\mathbb{E}[A_{t+\tau}^m \mid H_t^A]$ reduces to ordinary delay-coordinate prediction. Under stochastic dynamics, the conditional law generally remains broad, and different moments summarize different features of that law. 

Since the problem is merely a standard least-squares problem, the data-driven method for estimating the Koopman matrix is straightforward. For details, see \cite{Williams2015}. Here, we consider a situation where only one variable is observed; in the following discussion, we denote the variable as $A_t$.

Let $q=d_{\mathrm{E}}-1$ in \eqref{eq:delay_history}. Then, the history contains $d_{\mathrm{E}}$ coordinates, i.e.,  the current value and $d_{\mathrm{E}}-1$ delays. Define a polynomial lifting as
\begin{align}
 \bm{\psi}(H_t^A)
 =\left(1,A_t, A_{t-\tau}, A_{t-2\tau}, \dots, A_{t-(d_{\mathrm{E}}-1)\tau}, A_t^2, A_{t-\tau}^2, \dots,
A_{t-(d_{\mathrm{E}}-1)\tau}^K \right)^{\top},
 \label{eq:dictionary}
\end{align}
containing monomials of total degree not larger than $K$; according to \eqref{eq:mz_no_orthogonal}, we do not include cross-terms such as $A_t A_{t-\tau}$ in $\bm{\psi}(H_t^A)$. Then, the number of lifted features is 
\begin{align}
 N_{\mathrm{dic}} = 1 + d_{\mathrm{E} }K.
 \label{eq:dictionary_dimension}
\end{align}
Since the presence of a cross-term requires a total of $\binom{d_{\mathrm{E}}+K}{K}$ terms, $N_{\mathrm{dic}}$ becomes significantly smaller than when there is a cross-term. For training anchors $t \in \mathcal T_{\mathrm{train}}$, define the lifted snapshot matrices as follows:
\begin{align}
 \Psi_A=
 \begin{bmatrix}
 \bm{\psi}(H_{t_1}^A)^{\top}\\
 \vdots\\
 \bm{\psi}(H_{t_{N_{\mathrm{train}}}}^A)^{\top}
 \end{bmatrix},\qquad
 \widetilde{\Psi}_A=
 \begin{bmatrix}
 \bm{\psi}(H_{t_1+\tau}^A)^{\top}\\
 \vdots\\
 \bm{\psi}(H_{t_{N_{\mathrm{train}}}+\tau}^A)^{\top}
 \end{bmatrix}.
 \label{eq:edmd_data_matrices}
\end{align}
Then, ridge-regularized EDMD solves
\begin{align}
 \widehat{K}_A
 =\argmin_{\widetilde{K}} \left( \|\widetilde{\Psi}_A - \Psi_A \widetilde{K}^{\top}\|_F^2
 +\lambda\| \widetilde{K}\|_F^2 \right),
 \label{eq:edmd_ridge}
\end{align}
where $\| \cdot \|$ is the Frobenius norm, and $\lambda$ is a hyperparameter for the ridge-regularization. Hence, \eqref{eq:edmd_ridge} has the following closed-form solution:
\begin{align}
 \widehat{K}_A^{\top}
 =(\Psi_A^{\top} \Psi_A + \lambda I)^{-1}
 \Psi_A^{\top} \widetilde{\Psi}_A.
 \label{eq:edmd_closed_form}
\end{align}
In the infinite-data/increasing-dictionary idealization, EDMD is a Galerkin-type approximation of the Koopman operator; convergence and finite-dimensional approximation issues have been studied extensively \cite{Williams2015,Klus2016,Korda2018}. Here, we use EDMD only as a regression-based approximation to selected conditional moments.

Let $i^{(1)}$ be the index corresponding to $A_t$ and $i^{(2)}$ the index corresponding to $A_t^2$ in the dictionary vector $\bm{\psi}$ in \eqref{eq:dictionary}. Then, for example, the first and second moments are given with the estimated Koopman matrix $\widehat{K}_A$ as follows, respectively:
\begin{align}
 \mathbb{E}[A_{t+\tau} \mid H_t^A] 
= e_{i^{(1)}}^{\top}\widehat{K}_A\bm{\psi}(H_t^A),
 \qquad
 \mathbb{E}[(A_{t+\tau})^2 \mid H_t^A] 
= e_{i^{(2)}}^{\top}\widehat{K}_A \bm{\psi}(H_t^A).
 \label{eq:edmd_first_second}
\end{align}

\subsection{Numerical validation with additive and multiplicative coupled logistic maps}
\label{sec:koopman_validation}

We next see that the EDMD construction estimates the conditional moments associated with the chosen delay history. The two examples are designed so that the source $X$ influences $Y$ through two fundamentally different statistical channels. In the above discussions, we discretized continuous time using time lags of $\tau$. However, in the rest of the present paper, we will treat time as discrete, with time steps $t = 0, 1, 2, \dots$.

The system with additive coupling and that with multiplicative coupling are given as follows: 
\begin{align}
(\textrm{additive}) \quad  
&X_{t+1}=r_xX_t(1-X_t)+\sigma_x \, \xi_t^x, \quad
 Y_{t+1}=r_yY_t(1-Y_t)-cX_t+\sigma_y \, \xi_t^y,  \label{eq:additive_xy}  \\
(\textrm{multiplicative}) \quad
& X_{t+1}=r_xX_t(1-X_t)+\sigma_x \, \xi_t^x, \quad
 Y_{t+1}=r_yY_t(1-Y_t)+cX_t \,\xi_t^y, 
 \label{eq:multiplicative_xy}
\end{align}
where $\xi_t^x$ and $\xi_t^y$ are independent standard normal variables. Note that if the noise term is large, it may exceed the range of $[0,1]$. In such cases, we discard that trajectory and only use trajectories that do not exhibit divergence. For the additive coupling case, the structural direction is $X\to Y$ because $X_t$ enters the drift of $Y_{t+1}$ while $Y$ does not appear in the equation for $X$. For the multiplicative coupling case, there is also a structural direction $X\to Y$, but $X_t$ does not change the one-step conditional mean of $Y$; it modulates the conditional variance. The pair of models therefore provides a clean test of whether a method is estimating a full conditional law, only a first moment, or a quantity conditioned on a different information set.

\begin{figure}[bt]
\centering
\includegraphics[width=120mm]{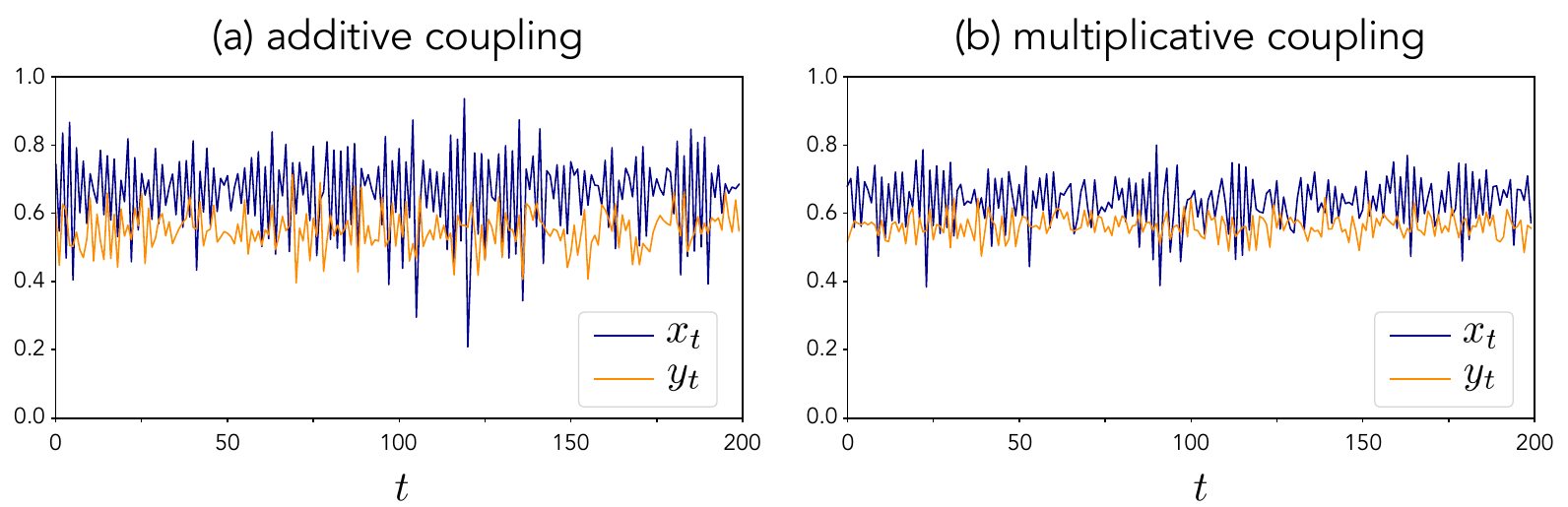}
\caption{Sample trajectories of the two stochastic coupled-logistic models used to validate the conditional-moment interpretation. (a) Additive coupling in \eqref{eq:additive_xy}, for which $X_t$ enters the conditional mean of $Y_{t+1}$. (b) Multiplicative coupling in ~\eqref{eq:multiplicative_xy}, for which $X_t$ modulates the amplitude of the noise term in $Y_{t+1}$. The figure contains $200$ samples from trajectories of total retained length $500{,}000$ after $500$ relaxation steps.}
\label{fig:xy_series}
\end{figure}

Figure~\ref{fig:xy_series} shows sample trajectories of the two stochastic coupled-logistic models. Here, $r_x=3.1$, $r_y=2.8$, $c=0.30$, $\sigma_x=\sigma_y=0.05$ for the additive coupling case, and $r_x=2.8$, $r_y=2.3$, $\sigma_x=0.05$, $c=0.05$ for the multiplicative coupling case. The results show that, despite having different coupling mechanisms, both models exhibit time series that, at first glance, appear to contain simple noise.

For each model, $500{,}000$ samples are retained after $500$ relaxation steps. To estimate the conditional expectations, the first $70\%$ of the trajectory forms a contiguous training block and the final $30\%$ a test block. We set the embedding dimension to $d_{\mathrm{E}} = 6$; a history uses the current value and five past values, i.e.,
\begin{align}
 H_t^A = (A_t, A_{t-1}, \dots, A_{t-5}), \qquad (A = X \textrm{ or } Y)
 \label{eq:history_six}
\end{align}
and the delay-space dimension is six. All monomials up to total degree $K=5$ are included and the ridge parameter is $\lambda=10^{-10}$. Note that the above settings yield sufficient accuracy in several preliminary numerical experiments. Hence, we adopt the above settings here.

To prevent information leakage, standardization is performed using the training block only. If $\mu_{A,\mathrm{train}}$ and $s_{A,\mathrm{train}}$ are the training mean and standard deviation of an observed variable, the lifted history uses
\begin{align}
 \widetilde{A}_t=\frac{A_t-\mu_{A,\mathrm{train}}}{s_{A,\mathrm{train}}},
 \label{eq:standardization}
\end{align}
and the same transformation is applied unchanged to the test block. Separate single-variable Koopman matrices are fitted for $X$ and $Y$; we obtain $\widehat{K}_X$ and $\widehat{K}_Y$ from the training dataset $\{X_t\}$ and $\{Y_t\}$, respectively. For each variable, the training pairs are $(H_t^A,H_{t+1}^A)$, so the complete lifted vector at the next time is regressed on the complete lifted vector at the current time. From the obtained Koopman matrices, we can evaluate the first- and second-moment predictions for the standardized dataset; we then extract the components corresponding to $A_{t+1}$ and $(A_{t+1})^2$ and transform them back to the original physical units.

We compare three calculations. The first is a one-step Monte Carlo (MC) benchmark with the original time-evolution equations in \eqref{eq:additive_xy} and \eqref{eq:multiplicative_xy}. A crucial point for the MC benchmark is that the MC simulation is initialized from the \emph{full current two-variable state} $(x_t, y_t)$. The MC mean therefore estimates
\begin{align}
 \mu^{\mathrm{MC}}_A(x_t,y_t)
\simeq \mathbb{E}[A_{t+1}\mid X_t=x_t,Y_t=y_t].
 \label{eq:mc_full_condition}
\end{align}
For each displayed anchor $(x_t,y_t)$, the code draws $N_{\mathrm{MC}}=1000$ independent one-step noise terms and evaluates
\begin{align}
 \widehat{\mu}_A^{\mathrm{MC}}=\frac{1}{N_{\mathrm{MC}}} \sum_{j=1}^{N_{\mathrm{MC}}} A_{t+1}^{(j)},
 \qquad
 \left(\widehat{\sigma}_A^{\mathrm{MC}}\right)^2
 = \frac{1}{N_{\mathrm{MC}}} \sum_{j=1}^{N_{\mathrm{MC}}}
 (A_{t+1}^{(j)}-\widehat{\mu}_A^{\mathrm{MC}})^2,
 \label{eq:mc_sample_moments}
\end{align}
where $\{A_{+1t}^{(j)}\}$ is the $j$-th independent one-step realization from the same current full state.

The second is the EDMD prediction in \eqref{eq:edmd_first_second}; we evaluate the first and second moments for the test datasets by using the Koopman matrices $\widehat{K}_X$ and $\widehat{K}_Y$, described above. Note that the expectation is different from the MC case; in the EDMD prediction, we evaluate $\mathbb{E}[A_{t+1} \mid H_t^A]$, i.e., the condition is based solely on the observed variable. In EDMD, the variance and standard deviation are calculated after obtaining the first and second moments. Although this operation does not theoretically guarantee that the variance is non-negative when using finite data, it was non-negative in all cases in our numerical experiments.

The third is a direct empirical conditional estimator based on $k=1000$ nearest training histories. For a test history $H_t^{A}$, let $\mathcal{N}^{(k)}(H_t^A)$ be the set of selected neighbor indices in the training dataset, and $d_i$ be the Euclidean distance to its $i$-th selected neighbor in $\mathcal{N}^{(k)} (H_t^A)$. In addition, set an adaptive bandwidth $b$ to the distance to the farthest data point within the cluster $\mathcal{N}^{(k)} (H_t^A)$. Then, the normalized Gaussian weights are
\begin{align}
 w_i(H_t^A)=
 \frac{\exp[-\frac{1}{2}(d_i/b)^2]}
 {\sum_{j\in \mathcal{N}^{(k)} (H_t^A)}\exp[-\frac{1}{2}(d_j/b)^2]}.
 \label{eq:neighbor_weights}
\end{align}
If $A_{t+1}^{(i)}$ is the one-step successor of the $i$-th training anchor in $\mathcal{N}^{(k)} (H_t^A)$, then
\begin{align}
 \widehat{\mu}_{A}^{\mathrm{NN}}=\sum_{i\in \mathcal{N}^{(k)} (H_t^A)} w_i A_{t+1}^{(i)},\qquad
 \left(\widehat{\sigma}_{A}^{\mathrm{NN}} \right)^2 = \sum_{i\in \mathcal{N}^{(k)} (H_t^A)} w_i (A_{t+1}^{(i)}-\widehat{\mu}_{A}^{\mathrm{NN}})^2.
 \label{eq:neighbor_moments}
\end{align}
This is an adaptive-bandwidth kernel regression of Nadaraya--Watson type \cite{Nadaraya1964,Watson1964}, with the $k$th-neighbor distance setting the local bandwidth. Equivalently, the weighted successors define an empirical approximation to $p(A_{t+1} \mid H_t^A)$. It therefore provides a nonparametric check on whether the polynomial EDMD model learns the same history-conditioned first and second moments.

\begin{figure}[t]
\centering
\includegraphics[width=150mm]{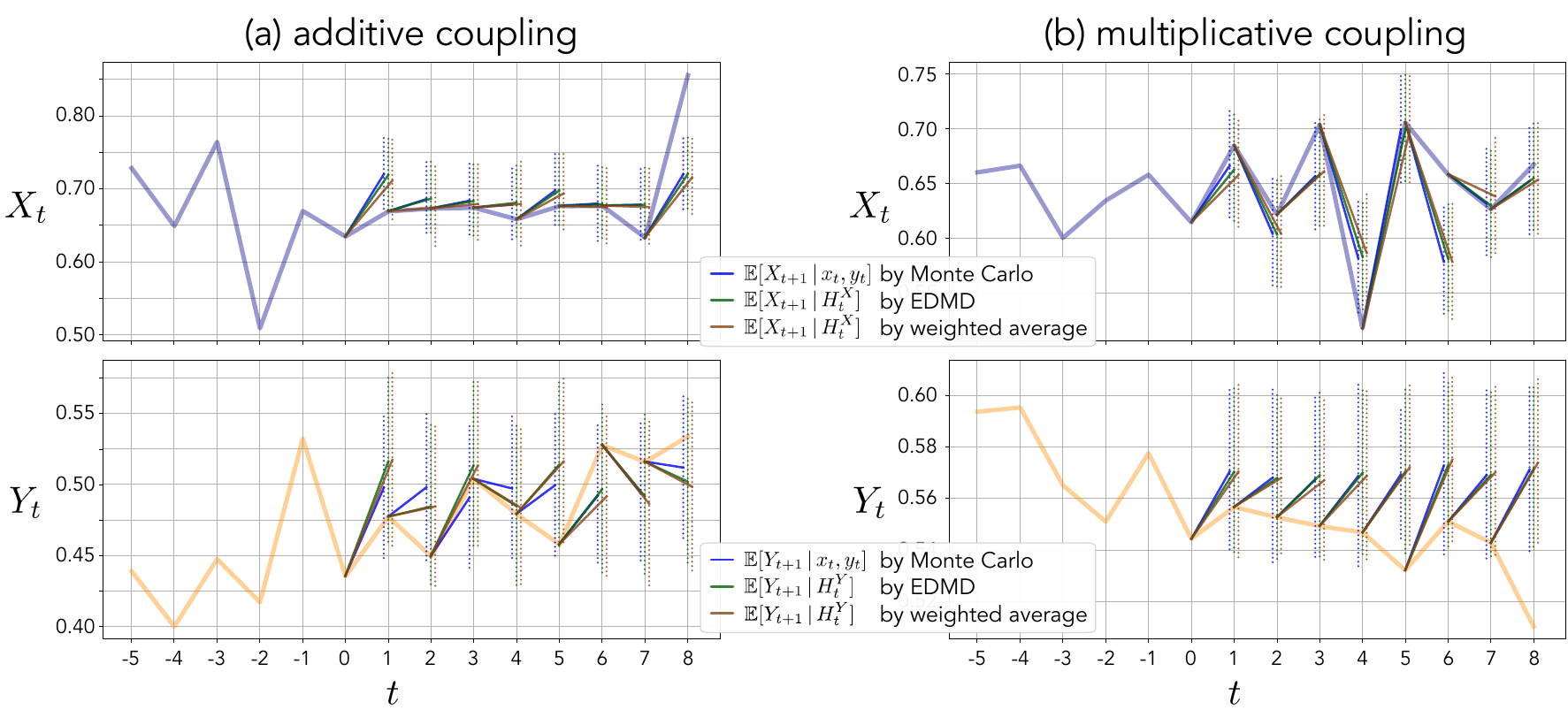}
\caption{Numerical results of one-step predictions in (a) additive and (b) multiplicative coupling cases. Upper and lower panels show predictions for $X$ and $Y$, respectively. Thick pale lines are test data. The three superposed prediction markers compare the three calculations: the Monte Carlo sampling initialized from the full current state $(X_t,Y_t)$, polynomial EDMD based on a single-variable six-dimensional history, and the weighted $k$-nearest-neighbor estimate based on the same history. Dotted vertical segments show one predicted standard deviation. The small horizontal offsets are only for visualization and do not represent temporal lags.}
\label{fig:koopman_check}
\end{figure}

Figure~\ref{fig:koopman_check} shows the prediction results. Although all the results appear to show good agreement, looking at the results for $Y_{t}$ in the additive coupling case (shown in the lower-left panel), there are parts where the MC results deviate significantly from the other two. As mentioned above, this is because the MC benchmark and the history-based estimators condition on different information sets. Below, we examine the reasons for each of the additive and multiplicative cases.

For the additive model, the MC estimates
\begin{align}
 \mathbb{E}[Y_{t+1} \mid X_t,Y_t]
 = f_y(Y_t)-cX_t,
 \qquad f_y(y)=r_yy(1-y),
 \label{eq:additive_fullmean}
\end{align}
whereas conditioning only on the observed $Y$ history gives
\begin{align}
 \mathbb{E}[Y_{t+1} \mid H_t^Y]
 = f_y(Y_t) - c \mathbb{E}[X_t\mid H_t^Y].
 \label{eq:additive_historymean}
\end{align}
The two expressions are equal only when $X_t$ is effectively known from $H_t^Y$ or when the particular realized $X_t$ happens to equal its conditional mean. Consequently, the MC estimates in the lower-left panel of figure~\ref{fig:koopman_check} are displaced from the EDMD and weighted-neighbor predictions.

For $X$, by contrast, the dynamics is autonomous in both examples. Hence
\begin{align}
 \mathbb{E}[X_{t+1} \mid X_t,Y_t]
 =\mathbb{E}[X_{t+1} \mid H_t^X]
 =r_xX_t(1-X_t),
 \label{eq:x_same_condition}
\end{align}
up to finite-sample estimation error, which explains the close agreement in the upper-left panel.

Next, we discuss why such a deviation does not occur in the case of multiplicative coupling. From \eqref{eq:multiplicative_xy}, independence and zero mean of $\xi_t^y$ imply
\begin{align}
 \mathbb{E}[Y_{t+1}\mid X_t,Y_t] = f_y(Y_t).
 \label{eq:multiplicative_fullmoments}
\end{align}
After conditioning only on the $Y$ history, we have
\begin{align}
 \mathbb{E}[Y_{t+1} \mid H_t^Y]=f_y(Y_t).
 \label{eq:multiplicative_historymoments}
\end{align}
Thus, the first conditional moment is the same under full-state and history-only conditioning even though the variances are different. This explains why the multiplicative case (the lower-right panel) shows much closer agreement of the predicted means than the additive case. 

These results are consistent with the discussion in \cite{Baldovin2020}. In \cite{Baldovin2020}, the effects of partial observation on the correlation functions are investigated in order to discuss causality. As a result, the authors in \cite{Baldovin2020} concluded that, in stochastic systems, partial observation affects the behavior of the correlation functions, and that inferring causality requires knowledge of variables relevant to the observed variables. As discussed so far, it is important to note that what can be computed via embedding in a stochastic system is, strictly speaking, only the conditional expectation conditioned on the embedding of the observed variables.

\section{CCM-like reconstruction by conditional expectations}
\label{sec:ccm}

To explore the potential applications of the embedding proposition for the stochastic systems discussed so far, we here address the problem of causal inference. One method of causal inference that utilizes the embedding theorem is CCM \cite{Sugihara2012}. However, CCM is theoretically rooted in deterministic state-space reconstruction, and process noise can degrade or alter its directional signatures. As discussed above, we extended the embedding argument to stochastic systems, and we therefore test a CCM-like method based on this extension.

\subsection{Review of convergent cross mapping}
\label{sec:ccm_review}

Granger causality, a well-known method for causal inference, defines direction operationally through improvement of prediction after adding a candidate source history \cite{Granger1969}. By contrast, CCM is rooted in deterministic state-space reconstruction: if $X$ drives $Y$, the reconstructed manifold of the response $Y$ can contain information about the source $X$, and one therefore maps from the response manifold to the putative source \cite{Deyle2011,Sugihara2012}. Lagged CCM uses the temporal displacement of the cross-map target to distinguish delayed influence from synchronization-like effects \cite{Ye2015,Deyle2016}. At the same time, theoretical and numerical studies have established important limitations; although state-space reconstruction may recover the transitive closure rather than only direct edges, the directionality could depend on the dynamical regime, and noise, synchronization, common drivers, sampling, and finite data can all produce misleading pairwise signatures \cite{McCracken2014,Cummins2015,Cobey2016,Monster2017,Runge2018,Runge2019}. These observations motivate an explicitly stochastic reformulation of what is reconstructed by a cross-map.

\subsubsection{Embedding and causality}

We first review CCM briefly \cite{Sugihara2012} because its arrow convention is sometimes confused with ordinary regression direction; the distinction becomes even more important after replacing deterministic state reconstruction by conditional expectation.

Let a deterministic coupled system yield scalar observations $X_t$ and $Y_t$. For embedding dimension $d_{\mathrm{E}}$ and delay $\tau$, the reconstructed manifold from $Y$ consists of points
\begin{align}
 \bm{Y}_t^{(d_{\mathrm{E}})}
 =(Y_t, Y_{t-\tau}, \dots, Y_{t-(d_{\mathrm{E}}-1)\tau}).
 \label{eq:y_manifold}
\end{align}
Under a valid deterministic embedding, there exists a one-to-one smooth correspondence between the original attractor state and the reconstructed point $\bm{Y}_t^{(d_{\mathrm{E}})}$. Consequently, any smooth observable of the full state, including $X_t$, is represented as a function on the $Y$ reconstructed manifold,
\begin{align}
 X_t = \widetilde{h}_{X|Y}\left(\bm{Y}_t^{(d_{\mathrm{E}})}\right).
 \label{eq:crossmap_function}
\end{align}
The cross-map is a data-driven local approximation to this function.

Suppose the physical interaction is $X \to Y$.  Because the evolution of $Y$ depends on $X$, the joint state information carried by $X$ is encoded in the sequence of $Y$ values. Hence, CCM tests the apparently reversed reconstruction
\begin{align}
 X \to Y \quad \Longrightarrow \quad \textrm{(reconstructed manifold by $Y$)} \to X.
 \label{eq:ccm_direction}
\end{align}
This reversal is fundamental: the history of the effect is used to reconstruct the cause. It is not the same as forecasting $Y$ from past $X$, as in a Granger-style predictive test.

\subsubsection{Simplex cross-map algorithm}

The library size $L$ is the number of embedded points available for constructing the local cross-map; prediction skill is then evaluated on a specified prediction set. The larger the number of data points, the more accurate the estimate will be. Under suitable deterministic embedding conditions, cross-map skill is expected to improve and approach a limiting value as the library samples the reconstructed manifold more densely. CCM utilizes this concept.

Choose a library of $L$ embedded points
\begin{align}
 \mathcal{L}_L = \left\{ \left(\bm{Y}_{t_i}^{(d_{\mathrm{E}})}, X_{t_i} \right) \right\}_{i=1}^{L}.
 \label{eq:ccm_library}
\end{align}
For a query $\bm{Y}_t^{(d_{\mathrm{E}})}$, locate its $d_{\mathrm{E}}+1$ nearest library neighbors, indexed by $t_1,\ldots,t_{d_{\mathrm{E}}+1}$ and ordered by distance. Define
\begin{align}
 d_i = \left\| \bm{Y}_t^{(d_{\mathrm{E}})} - \bm{Y}_{t_i}^{(d_{\mathrm{E}})} \right\|_2,
 \qquad
 u_i = \exp\left[-\frac{d_i}{d_1+\epsilon}\right],
 \qquad
 w_i = \frac{u_i}{\sum_{j=1}^{d_{\mathrm{E}}+1}u_j},
 \label{eq:ccm_weights}
\end{align}
and estimate
\begin{align}
 \widehat{X}_t^{\mathrm{CCM}}
 =\sum_{i=1}^{d_{\mathrm{E}}+1} w_i X_{t_i}.
 \label{eq:ccm_estimator}
\end{align}
The use of $d_{\mathrm{E}}+1$ neighbors constructs the local simplex in the reconstruction. Increasing $d_{\mathrm{E}}$ is therefore not the same operation as increasing the number of neighbors: $d_{\mathrm{E}}$ changes the reconstructed coordinate space, while the neighbor rule approximates a local map within that chosen space.

Cross-map skill is commonly measured by Pearson correlation,
\begin{align}
 \rho_{X \leftarrow M_Y}(L)
 =\Corr(X_t, \widehat{X}_t^{\mathrm{CCM}}),
 \label{eq:ccm_rho}
\end{align}
or by a related prediction-error statistic. In CCM, we repeat the calculation for increasing library sizes $L$. A small library sparsely samples the attractor; nearest neighbors can therefore be far apart in true state even when the embedding coordinates are formally valid. As $L$ grows, local neighborhoods become more densely sampled and, in an ideal deterministic setting, the cross-map estimate should approach the underlying function in \eqref{eq:crossmap_function} \cite{Sugihara2012}.

The roles of three quantities should be kept separate:
\begin{align*}
\begin{aligned}
 d_{\mathrm{E}} &:\quad \text{dimension of each delay-coordinate},\\
 \tau &:\quad \text{temporal spacing for observation},\\
 L &:\quad \text{number of library points}.
\end{aligned}
\end{align*}
Changing $d_{\mathrm{E}}$ changes what information is supplied to the local map; changing $L$ changes how densely that map is learned. Merely increasing $d_{\mathrm{E}}$ is not a substitute for library-size convergence.

Again, we will consider a discrete-time system with $t = 0,1,2,\dots$ in the following discussions and assume that $\tau = 1$ for simplicity.

\subsection{Time-shifted conditional expectations and the meaning of library size}
\label{sec:shifted_conditional}

As discussed above, in stochastic systems, we need to consider conditional expectations rather than the values themselves. For a stochastic system, a small neighborhood of an observed delay vector generally corresponds to a distribution of compatible hidden states rather than a unique state. Therefore, the direct stochastic analogue of the deterministic cross-map function~\eqref{eq:crossmap_function} is a conditional statistic. For a predictor channel $A$, define
\begin{align}
 H_t^A = (A_t,A_{t-1}, \dots, A_{t-(d_{\mathrm{E}}-1)})
 \label{eq:ccm_history}
\end{align}
and let the target channel be $B$. Then, for an integer lag $\ell$ and moment order $m$, we investigate $\mathbb{E}[B_{t-\ell}^{m} \mid H_t^A]$. In particular, the following discussions and numerical experiments focus solely on $m = 1$ as a first step.

Why should we use a shifted target at all? In the deterministic CCM algorithm, if $B$ contributes causally to the generation of $A$, then present response coordinates can contain information about earlier source states. A causal imprint is therefore expected to be asymmetric in time. Conversely, purely simultaneous correlation, synchronization, or common external forcing can generate high same-time reconstructibility without the temporal ordering expected from a directed interaction. Scanning $\ell$ asks where the information carried by $H_t^A$ is strongest in the target's temporal neighborhood. This is closely related in spirit to extended/lagged CCM \cite{Ye2015}, but here the reconstructed object is explicitly a conditional moment.

\begin{figure}[t]
\centering
\includegraphics[width=120mm]{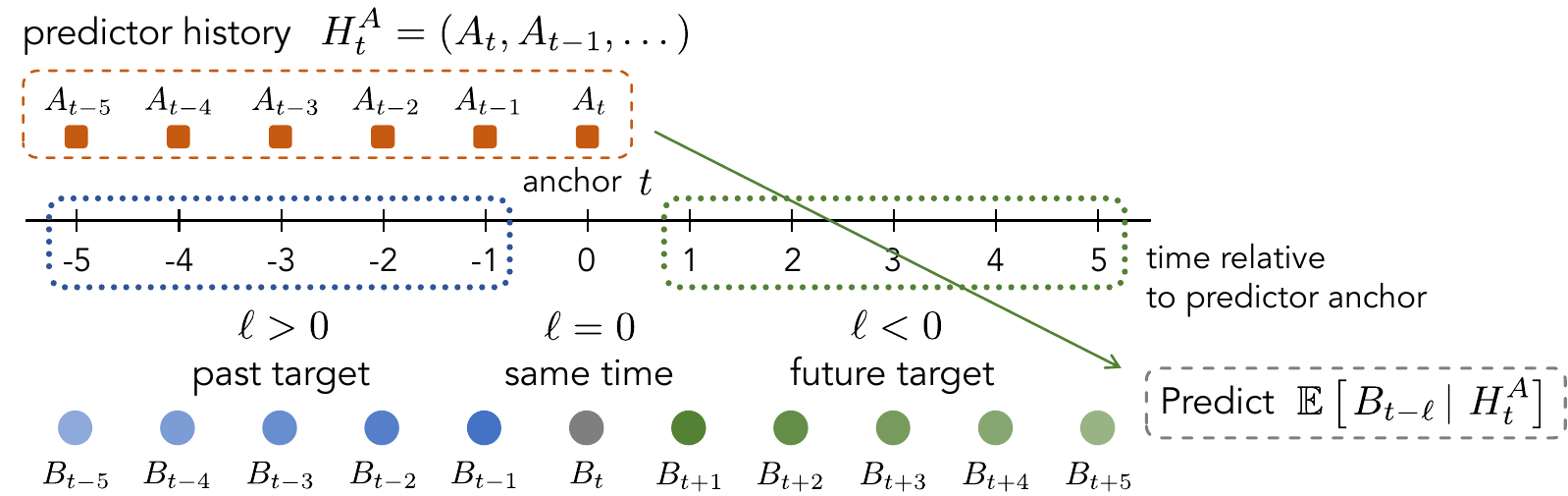}
\caption{Definition of the time-shifted conditional reconstruction. The predictor history is $H_t^A=(A_t,A_{t-1},\dots)$.  For lag $\ell$, the target is $B_{t-\ell}$. Hence $\ell>0$ reconstructs the target's past, $\ell=0$ the simultaneous target, and $\ell<0$ the target's future.  The regression approximates $\mathbb{E}[B_{t-\ell} \mid H_t^A]$, and the correlation between the conditional-expectation estimate and the realized shifted target defines the reported reconstruction skill.}
\label{fig:lag_scheme}
\end{figure}

Our lag convention is shown in figure~\ref{fig:lag_scheme}:
\begin{align}
 \ell>0:\ B_{t-\ell}\ \text{is in the past},\qquad
 \ell=0:\ B_t,\qquad
 \ell<0:\ B_{t-\ell}\ \text{is in the future}.
 \label{eq:lag_sign_convention}
\end{align}
Because some lagged-CCM papers adopt the opposite sign convention, equations rather than verbal labels should be compared across studies.

The method for calculating the conditional expected value is the same as that described in section~\ref{sec:discrete_condmom}. While it is possible to use a weighted average, we employ ridge regression that is linear in a polynomially lifted feature space in the following numerical experiments. Note that, in section~\ref{sec:discrete_condmom}, we predicted the next $A$ based on $A$'s current-and-past delay history. By contrast, we here predict the expected value of $B$ based on $A$'s current-and-past delay history.

Given a library of $L$ training anchors, a regression model $g_{L,\ell}^{(m)}$ is fitted to approximate
\begin{align}
 \widehat{\mu}_{B|A,L}^{(m)}(\ell;H_t^A)
 =g_{L,\ell}^{(m)}(H_t^A)
 \simeq\mathbb{E}[B_{t-\ell}^{m} \mid H_t^A].
 \label{eq:estimated_lag_conditional}
\end{align}
Note that the library size $L$ corresponds to the number of training examples to construct the regression model. The score is evaluated on a fixed held-out block,
\begin{align}
 \rho_{B|A}^{(m)}(\ell; L)
 =\Corr\left(
 B_{t-\ell}^{m},
 \widehat{\mu}_{B|A,L}^{(m)}(\ell;H_t^A)
 \right).
 \label{eq:rho_general}
\end{align}
For the sake of simplicity, $\rho_{L}$ will be used to denote the correlation in \eqref{eq:rho_general} when it is clear from the context. 

The polynomial ridge-regression model used below differs computationally from the $(d_{\mathrm{E}}+1)$-neighbor simplex interpolator, but the cross-map direction is kept identical to CCM. Throughout this paper, the hypothesis $X\to Y$ is tested by the reconstruction $H_t^Y \longmapsto X_{t-\ell}$. That is, the history of the response is used to reconstruct the putative source. The reverse hypothesis $Y\to X$ uses $H_t^X$ to reconstruct $Y_{t-\ell}$.

The library size $L$ is the number of training anchors used to estimate the conditional map. In the implementation in the following numerical experiments, the first $L$ valid anchors of the contiguous training block are used; the test block is held fixed for all $L$. For each $L$, the predictor channel is standardized, a degree-five polynomial lift of the six-coordinate history is constructed, and ridge regression is fitted to the shifted raw target. Thus, an increase in $\rho_L$ reflects improved estimation of the same conditional regression function rather than a change of test data.

Note that, in a stochastic system, even an infinite library does not imply perfect pointwise reconstruction. Conditional expectation gives the orthogonal decomposition
\begin{align}
 \mathbb{E}\left[\bigl(B_{t-\ell}^m-g(H_t^A)\bigr)^2\right]
 =\mathbb{E}\left[\Var(B_{t-\ell}^m \mid H_t^A)\right]
 +\mathbb{E}\left[\bigl(\mu_{B|A}^{(m)}(\ell, H_t^A)-g(H_t^A)\bigr)^2\right].
 \label{eq:conditional_variance_floor}
\end{align}
The first term is irreducible uncertainty left after conditioning on the chosen history; only the second term is the estimation error. Consequently, stochastic ``convergence'' should mean convergence to the best conditional predictor allowed by the information set and function class, not convergence of $\rho_L$ to unity. This distinction is important to the numerical experiment below.

\subsubsection{Library-size test in additively coupled logistic maps}
\label{sec:library_test}

We now test whether library-size convergence alone reliably reveals the known direction $X \to Y$. The additive model has the same form as \eqref{eq:additive_xy}, but here we use $c=0.1$ and $\sigma_x=\sigma_y=0.04$.  Two parameter choices exchange the intrinsic logistic-map regimes:
\begin{align}
 \text{case (a)}:&\quad r_x=3.1,\quad r_y=2.8,\\
 \text{case (b)}:&\quad r_x=2.8,\quad r_y=3.1.
 \label{eq:library_cases}
\end{align}
For each case, $500{,}000$ points are retained after $500$ relaxation steps, the first $70\%$ are used for training, and the final $30\%$ are held out. Histories contain the current point and five delays ($d_{\mathrm{E}} = 6$), polynomial degree is $K=5$, and ridge strength is $10^{-10}$. For the numerical experiments, we repeat the generation and analysis of time-series data $10$ times, plotting the resulting mean and standard deviation. For each value of $L$, the first $L$ admissible anchors of the contiguous training block are used, while the test block is kept fixed, as denoted above. We use
\begin{align}
 L \in \{100, 500,1000,2000,5000,10000,20000,50000\}.
 \label{eq:library_sizes}
\end{align}
For the $X\to Y$ hypothesis, a polynomial map is fitted from $H_t^Y$ to $X_{t-1}$; the reverse direction fits $H_t^X$ to $Y_{t-1}$. Hence, this experiment fixes $\ell=1$ and varies only $L$.

\begin{figure}[t]
\centering
\includegraphics[width=120mm]{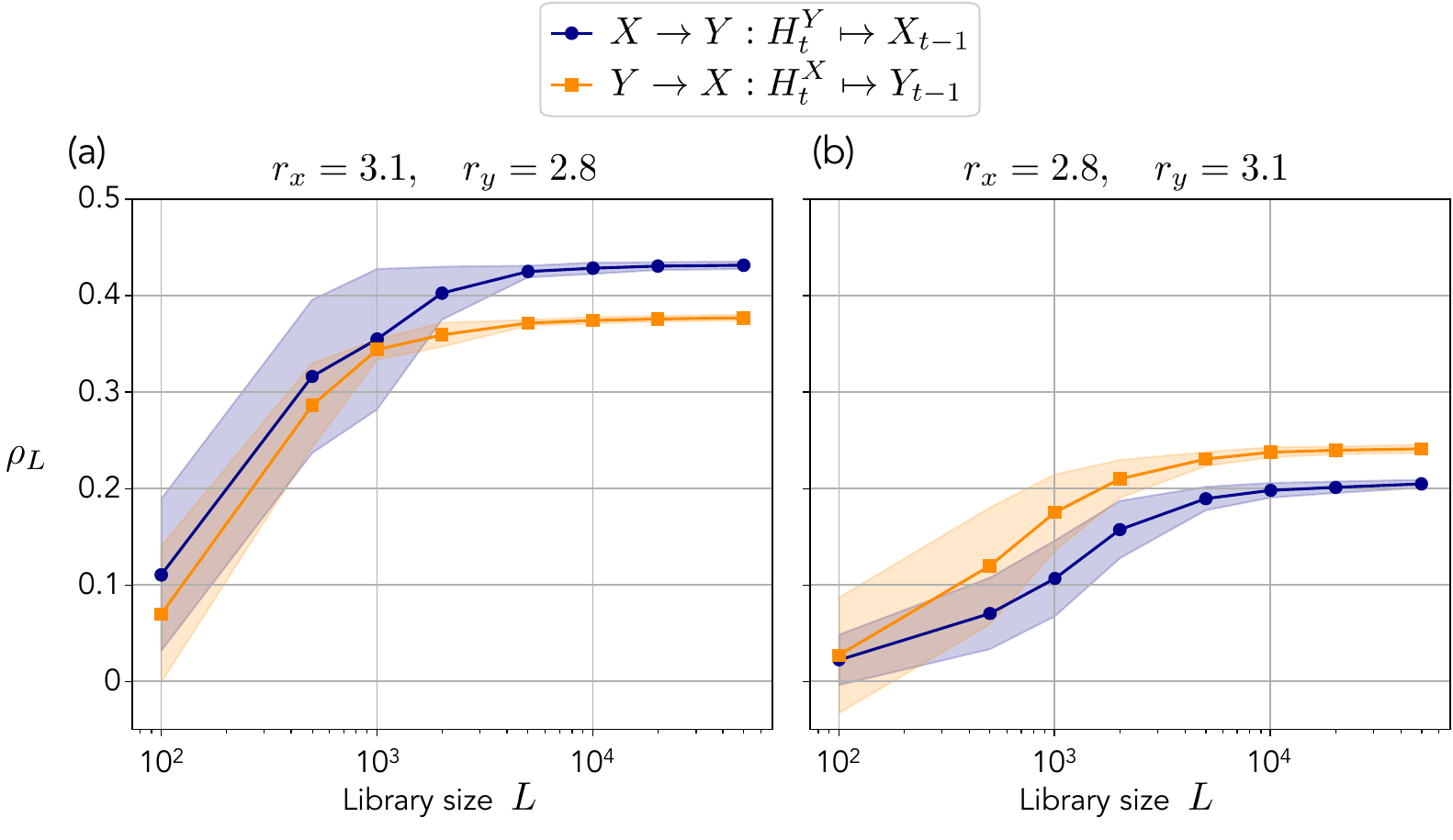}
\caption{Library-size dependence of the CCM-like first-moment reconstruction for the additive $X\to Y$ model.  The tested $X\to Y$ direction uses $H_t^Y \mapsto X_{t-1}$; the reverse uses $H_t^X\mapsto Y_{t-1}$.  (a) $r_x=3.1$, $r_y=2.8$: the correct-direction skill grows more strongly, but the reverse skill also increases. (b) $r_x=2.8$, $r_y=3.1$: both directions increase and, over much of the range, the reverse reconstruction is larger despite the unchanged structural direction $X\to Y$. Hence, convergence with $L$ by itself is not sufficient to identify direction in this stochastic example. The graph plots the mean and standard deviation of the results obtained from generating and analyzing $10$ individual time-series datasets.}
\label{fig:library}
\end{figure}

Figure~\ref{fig:library} shows the library-size dependence of the CCM-like first-moment reconstruction for the additive coupling system in \eqref{eq:additive_xy}. Note that the system parameters differ in (a) and (b) of figure~\ref{fig:library}. 

In case (a), the source $X$ has the stronger alternating deterministic structure associated with $r_x=3.1$, and this structure is injected into $Y$ through the additive drift. Consequently, the $Y$ history contains recoverable information about past $X$, and the nominal $X\to Y$ score rises substantially with $L$. The reverse score nevertheless also rises because $X$ and $Y$ are dynamically coherent, autocorrelated signals rather than independent samples.

In case (b), there is also a tendency for $\rho_{L}$ to increase as $L$ increases. However, the parameter exchange in \eqref{eq:library_cases} is important because the uncoupled logistic map $x_{t+1}=rx_t(1-x_t)$ changes its asymptotic regime as $r$ varies \cite{May1976}. Values near $r=2.8$ are strongly contracting toward a stable low-period behavior, whereas $r=3.1$ lies beyond the first period-doubling threshold and produces a richer oscillatory response. With additive process noise the invariant distributions are broadened, but the different local contraction rates remain relevant. A history generated by the more structured response is easier to regress on than one generated by the more weakly contracting or higher-period source, so the magnitude of a cross-map correlation is not determined by graph direction alone. This is precisely the dynamical-regime dependence emphasized in critiques and theoretical analyses of reconstruction-based causality \cite{McCracken2014,Cummins2015,Monster2017}.

\subsubsection{Lag-dependent conditional-moment causal diagnostics}
\label{sec:lagcausality}

Next, we investigate the lag dependence of the correlations. Under the convention of figure~\ref{fig:lag_scheme}, a genuinely causal response at time $t$ can contain information about a source at earlier times, so enhanced reconstructibility for $\ell>0$ is physically compatible with forward causation.  A large value only for $\ell<0$ means that the present predictor history best reconstructs a future target and is more naturally attributed to temporal predictability, synchronization, or a reversed relation than to a source-to-response delay.

A useful summary is the positive-lag location
\begin{align}
 \ell_{B\to A}^{\star}
 =\argmax_{\ell>0}\rho_{B|A}^{(m)}(\ell;L).
 \label{eq:lag_peak}
\end{align}
Note that this quantity is not automatically identified with the physical coupling delay. The history $H_t^A$ is itself a finite time window. If the source at $t-\ell$ influences several subsequent response coordinates that all lie inside the history window, then older source times are easier to reconstruct because the regression sees several descendants of the same perturbation. The position of the peak can consequently move when the embedding depth is changed. Hence, the lag profile is best viewed as an \emph{effective information-age profile} associated with a specified history length, dictionary, and moment order.

\begin{figure}[t]
\centering
\includegraphics[width=120mm]{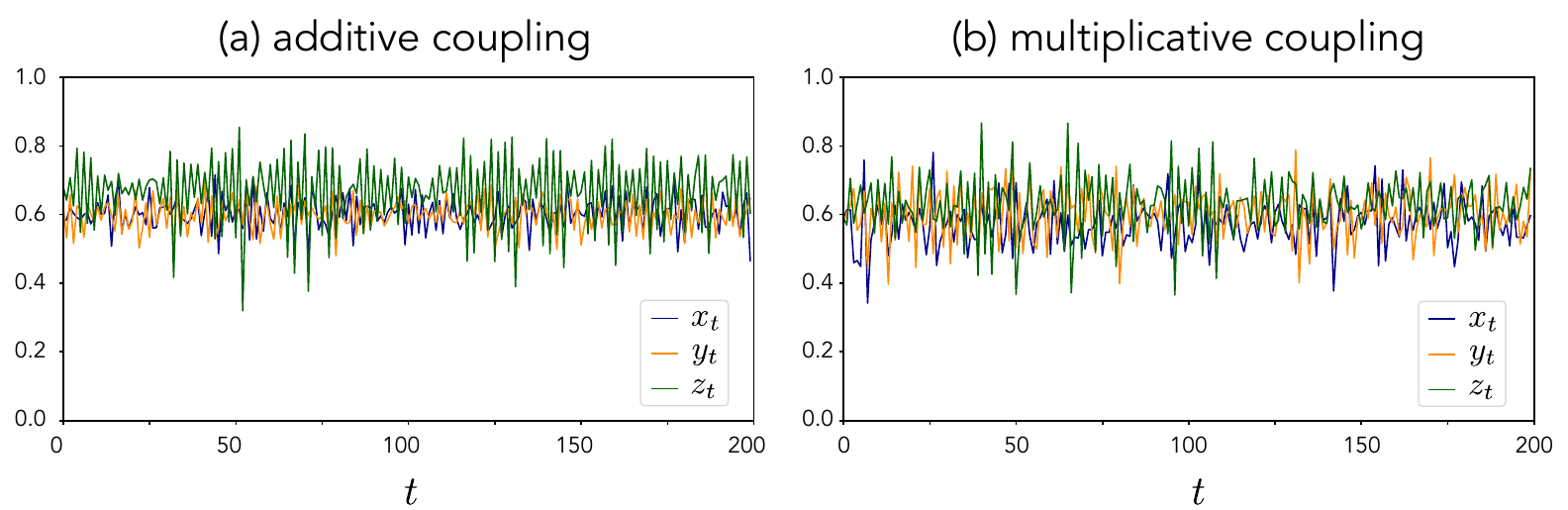}
\caption{Sample trajectories for the common-driver systems with structural graph $Z\to X$ and $Z\to Y$. (a) Additive driver coupling in  \eqref{eq:common_additive}. (b) Multiplicative driver coupling in \eqref{eq:common_multiplicative}. Although the pair $(X,Y)$ has no direct edge in either model, the two series share information through $Z$, providing a test of whether a pairwise lag diagnostic can distinguish common-driver structure from direct driver--response structure.}
\label{fig:common_series}
\end{figure}

In the following numerical experiments, we use examples to investigate the feasibility of extracting more complex causal relationships. There are three variables with the structural graph $Z\to X$ and $Z\to Y$, but no direct $X\leftrightarrow Y$ edge. The first system has additive driver coupling,
\begin{align}
\begin{cases}
\,\,  X_{t+1}=r_xX_t(1-X_t)-c_xZ_t + \sigma_x \xi_t^x,\\
\,\,  Y_{t+1}=r_yY_t(1-Y_t)-c_yZ_t + \sigma_y \xi_t^y,\\
\,\,  Z_{t+1}=r_zZ_t(1-Z_t)+\sigma_z\xi_t^z,
\end{cases}
 \label{eq:common_additive}
\end{align}
with $r_x=r_y=2.8$, $r_z=3.1$, $c_x=c_y=0.1$, and $\sigma_x=\sigma_y=\sigma_z=0.04$. The second system uses multiplicative driver coupling,
\begin{align}
\begin{cases}
\,\,  X_{t+1}=r_xX_t(1-X_t)+c_xZ_t\xi_t^x,\\
\,\,  Y_{t+1}=r_yY_t(1-Y_t)+c_yZ_t\xi_t^y,\\
\,\,  Z_{t+1}=r_zZ_t(1-Z_t)+\sigma_z\xi_t^z,
 \label{eq:common_multiplicative}
\end{cases}
\end{align}
with $ r_x=2.4$, $r_y=2.6$, $r_z=2.8$, $c_x=c_y=0.1$, and $\sigma_z=0.06$. Sample trajectories are shown in figure~\ref{fig:common_series}.

We fix the library size as $L=10{,}000$ and investigate the correlations for the lag range $-10\leq\ell\leq10$. Other parameter settings are the same as in the previous numerical experiments.

\begin{figure}[t]
\centering
\includegraphics[width=120mm]{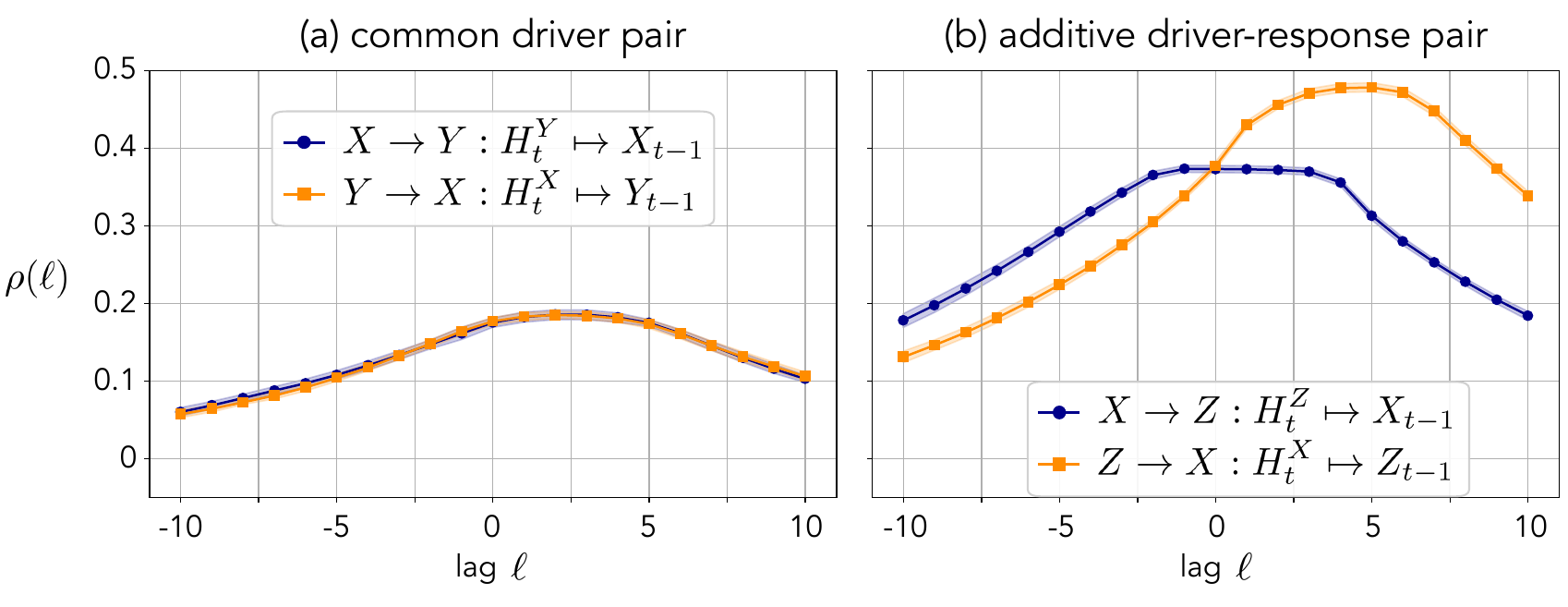}
\caption{Lag dependence of first-moment conditional reconstruction in the additive common-driver system. $\ell>0$ means the reconstruction of the target's past from the predictor history. (a) Common-driver pair $X$--$Y$.  The two directions have similar broad profiles because both variables are driven by $Z$ and there is no direct edge between them. (b) Driver--response pair $X$--$Z$. The structural direction $Z\to X$ is evaluated by $H_t^X \mapsto Z_{t-\ell}$ and is markedly stronger, with a maximum near $\ell=5$ for the six-coordinate history used here. The peak location is embedding dependent and is not interpreted as a five-step physical coupling delay. The graph plots the mean and standard deviation of the results obtained from generating and analyzing $10$ individual time-series datasets.}
\label{fig:additive_lag}
\end{figure}

Figure~\ref{fig:additive_lag} shows the first-moment lag scans for \eqref{eq:common_additive}. For the pair $(X,Y)$ in panel (a), the two curves are similar to each other and have broad shapes. This is the expected qualitative signature of the common driver: each response contains information about $Z$ and therefore about statistical structure shared with the other response, but neither equation contains a direct $X\to Y$ or $Y\to X$ term. A pairwise method can consequently find cross-predictability in both directions. The result is a warning that a lag scan is a diagnostic rather than a substitute for conditioning on measured common causes.

For the $(X,Z)$ pair in panel (b) in figure~\ref{fig:additive_lag}, the structural direction is $Z\to X$. In CCM arrow convention, this direction is tested by using $H_t^X$ to reconstruct $Z_{t-\ell}$. The orange curve is substantially larger than the reverse curve and, for the present six-coordinate history with five delays, reaches its maximum at $\ell=5$. The reverse $X\to Z$ curve is broad and weaker, and its peak is far from $\ell=5$. Thus, the lag dependence contains a much clearer directional signature for the actual driver--response pair than for the two variables linked only through a common driver.

Why does the maximum occur around $\ell=5$ even though \eqref{eq:common_additive} contains only a one-step coupling? Write the response schematically as
\begin{align}
 X_{t+1}=f_x(X_t) - c_x Z_t+\epsilon_t,
 \label{eq:additive_imprint}
\end{align}
where $\epsilon_t$ corresponds to a schematic noise. A target $Z_{t-1}$ has only one immediately available response descendant inside $H_t^X$, namely $X_t$. By contrast, $Z_{t-5}$ first affects $X_{t-4}$, and its direct and dynamically propagated consequences could also contribute to $X_{t-3}, \dots,X_t$. Hence, the six-coordinate history contains multiple post-influence observations associated with an older target. The regression combines those descendants to separate the driver signal from the self-dynamics and innovations of $X$. Increasing the history depth creates room for still older target times to have multiple descendants in the predictor window; accordingly, the empirical location of the maximum moves toward larger positive $\ell$ when the embedding depth is increased. This is precisely why $\ell^\star_{Z\to X}$ should be interpreted jointly with the chosen history length.

\begin{figure}[t]
\centering
\includegraphics[width=120mm]{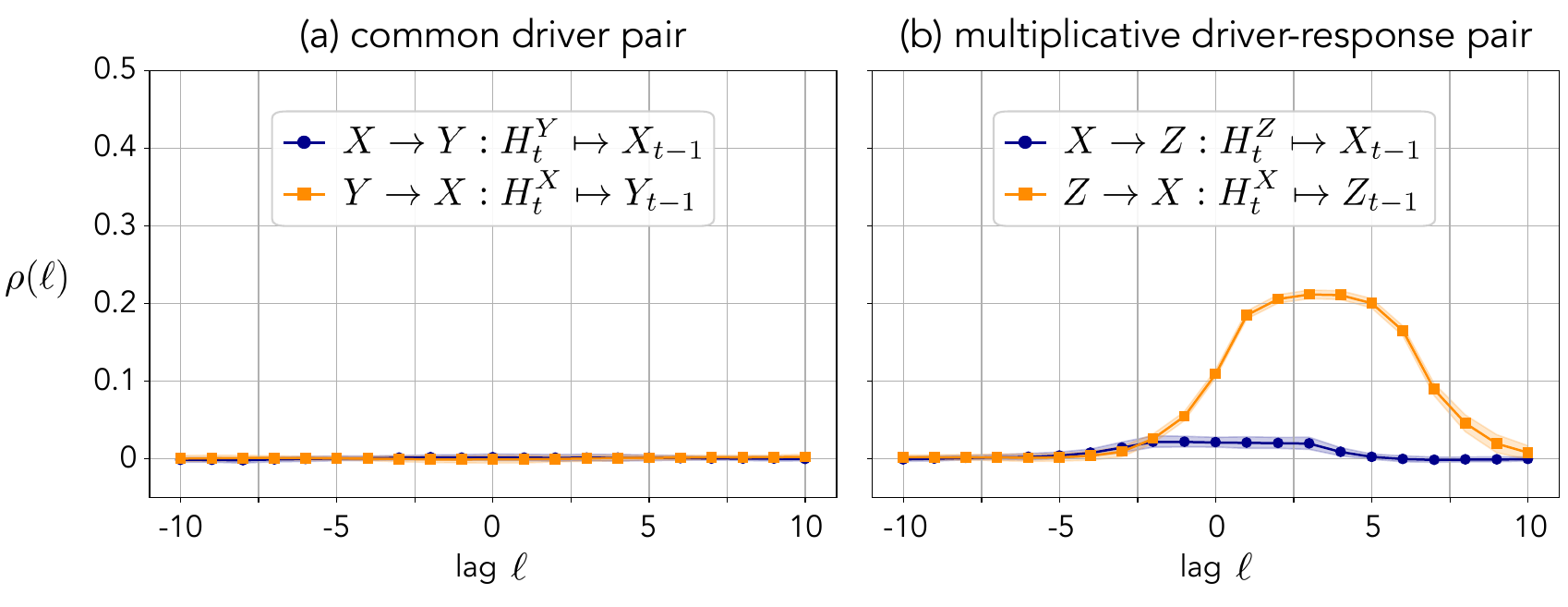}
\caption{Lag dependence of first-moment conditional reconstruction in the multiplicative common-driver system. (a) Common-driver pair $X$--$Y$: both curves remain near zero because the common driver enters the two response equations through independent zero-mean multiplicative innovations, so the leading one-step information is in conditional variances, not means. (b) Driver--response pair $X$--$Z$: the structural $Z\to X$ direction is visible indirectly in the first moment and peaks near $\ell=3$ or $\ell=4$. The signal arises after nonlinear propagation converts the $Z$-dependent variance of $X_{t+1}$ into a $Z^2$-dependent mean at later times; \eqref{eq:variance_to_mean} gives the simplest analytic illustration. The graph plots the mean and standard deviation of the results obtained from generating and analyzing $10$ individual time-series datasets.} 
\label{fig:multiplicative_lag}
\end{figure}

Figure~\ref{fig:multiplicative_lag} for the multiplicative case is qualitatively different. For the common-driver pair $(X,Y)$, both first-moment curves are essentially zero. By contrast, the driver--response pair $(Z,X)$ shows a nonzero $Z\to X$ first-moment lag profile, peaking near $\ell=3$ or $\ell=4$ for the present settings. This seemingly surprising signal is generated by nonlinear conversion of variance information into later mean information. Let $f_x(x) = r_x x (1-x)$ and condition on $(X_t,Z_t)$. We see that \eqref{eq:common_multiplicative} gives
\begin{align}
 \mathbb{E}[X_{t+1} \mid X_t,Z_t] = f_x(X_t),
 \qquad
 \mathbb{E}[X_{t+1}^2 \mid X_t,Z_t] = f_x(X_t)^2 + c_x^2 Z_t^2,
 \label{eq:mult_first_second}
\end{align}
and hence, the first conditional moment does not depend on $Z_t$. However, at the next step,
\begin{align}
 \mathbb{E}[X_{t+2}\mid X_t,Z_t]
 =r_x \mathbb{E}[X_{t+1}-X_{t+1}^2 \mid X_t,Z_t]
 =r_x \left[f_x(X_t)-f_x(X_t)^2 - c_x^2Z_t^2\right],
 \label{eq:variance_to_mean}
\end{align}
ignoring variables independent of the conditioning set that do not alter the displayed dependence. Thus, $Z_t^2$ appears in the two-step conditional mean because the nonlinear logistic drift acts on a random $X_{t+1}$. A degree-five polynomial dictionary can represent such quadratic dependence. Although these equations do not directly equal the history-conditioned cross-map, they demonstrate a mechanism by which $Z$-dependent variance affects later conditional means. That is, the peak in figure~\ref{fig:multiplicative_lag}(b) appearing at an $\ell$ smaller than that of the peak in figure~\ref{fig:additive_lag}(b) suggests that this effect is secondary.

Finally, we comment on the difference from the standard CCM. In the standard CCM, it is common to set the number of neighbors to $d_{\mathrm{E}}+1$, where $d_{\mathrm{E}}$ is the embedding dimension \cite{Sugihara2012}. In the preliminary numerical experiments, running the CCM with $d_{\mathrm{E}} = 6$ yielded results showing similar trends in the above approach. Additionally, we observed a tendency for correlation values to increase as the number of neighbors was increased. This is because the weighted calculations in CCM correspond to taking conditional expectations when the number of points in the neighborhood is large enough. In previous studies of deterministic systems, increasing the number of neighbors has been used to reduce the effects of noise. However, a crucial point is that the present work clarifies that, in stochastic systems, one should consider the conditional expectations. Consequently, it is expected that future research will shift toward increasing the number of neighbors not merely to average out the effects of noise, but specifically to improve the accuracy of the conditional expectations.

\section{Summary and future work}
\label{sec:summary}

We developed a stochastic interpretation of delay-coordinate reconstruction in which an observed history is used to reconstruct conditional moments rather than a unique hidden or future sample-path state. The viewpoint is supported by two complementary arguments. At the density level, an SDE generates deterministic Fokker--Planck evolution, and one can represent moment-determinate distributions by an infinite deterministic moment hierarchy. A finite closure gives a heuristic route back to the setting of ordinary deterministic delay embedding, although a rigorous stochastic Takens proposition is explicitly beyond the scope of the present paper. At the observable-function level, the Koopman operator propagates coordinates and their powers into conditional moments. The Mori--Zwanzig projection method further shows why a coefficient-space orthogonal term vanishes for a coordinate observable without eliminating stochastic uncertainty in the state process. In addition, the numerical conditional-moment test clarified the distinction between information sets.

As an application, we investigated causal diagnostics based on a CCM-like conditional-moment reconstruction in stochastic systems. For causal inference, we retained the CCM principle that the history of a variable affected by the cause is used to reconstruct its putative cause; we replaced deterministic point reconstruction by conditional-moment regression. The library-size experiment showed that increasing $\rho_L$ alone is not a reliable directional criterion for the noisy logistic examples. That is, the result depends strongly on the intrinsic dynamical regimes of the two variables. Then, we discussed that a lag scan adds temporal information. In the additive common-driver model, the two response variables showed similar weak pairwise curves while the true driver--response curve was markedly stronger. In the multiplicative model, the first-moment common-driver curves vanished almost completely, consistent with the fact that coupling acts through conditional variance.

Several limitations suggest directions for future research. First, more practical approaches to causal inference will be necessary. For example, polynomial dictionaries are convenient, but kernel methods or other universal approximators should be compared with explicit complexity control. Furthermore, the present work merely uses causal inference as a demonstration of the embedding proposition in stochastic systems. Hence, it is necessary to investigate its effectiveness in a wider range of applications. Second, the moment-based Takens discussion in section~\ref{sec:takens_moments} is intentionally heuristic. Establishing precise relations among finite moment closures, measure-theoretic embeddings, and Koopman--Mori--Zwanzig memory remains a mathematical problem in its own right.

Discussions regarding embedding propositions in stochastic systems are expected to broaden the scope of application for various methods related to the embedding theorem in deterministic systems. The present work represents a step toward connecting stochastic delay-coordinate reconstruction, Koopman--Mori--Zwanzig formalism, and causal diagnostics. As mentioned above, further studies are needed to derive practical causal inference methods from a statistical viewpoint. We hope this study will contribute to future research.

\begin{acknowledgments}
This work was supported in part by JST FOREST Program (Grant Number JPMJFR216K, Japan).
\end{acknowledgments}

\end{document}